\documentclass[10pt,twocolumn,letterpaper]{article}

\usepackage[pagenumbers]{cvpr} 

\usepackage{xspace}
\newcommand{\model}{CodeV\xspace}
\newcommand{\algo}{TAPO\xspace}
\newcommand{\tsc}[1]{\textsuperscript{#1}}

\usepackage{graphicx}
\usepackage{xcolor}
\usepackage{caption}
\usepackage{tcolorbox}
\tcbuselibrary{skins,breakable,listings}
\usepackage{listings}
\usepackage{inconsolata} 
\usepackage[T1]{fontenc}       
\usepackage{inconsolata}       

\usepackage{multirow}
\usepackage{booktabs}
\usepackage{makecell}
\usepackage{float}
\usepackage{algorithm}
\usepackage{algpseudocode}
\usepackage{fontawesome5}
\definecolor{thinkbg}{HTML}{E7F5FF}  
\definecolor{darkblue}{HTML}{bac8ff}
\definecolor{darkyellow}{HTML}{fff9db}
\newtcolorbox[auto counter,number within=section]{prompttemplate}[2][]{
  colback=gray!10,
  colframe=black!70,
  boxrule=0.6pt,
  arc=4pt,
  left=6pt,right=6pt,top=6pt,bottom=6pt,
  title={Template~\thetcbcounter: #2},
  #1
}

\newtcolorbox{myframe}{
  colback=white,
  colframe=black,
  arc=6pt,              
  boxrule=0.6pt,        
  left=3pt,
  right=3pt,
  top=3pt,
  bottom=3pt,
  enhanced
}
\newtcolorbox{promptbox}[1][]{
  enhanced,
  breakable,
  colback=white,          
  colframe=black!25,
  colbacktitle=black!8,   
  coltitle=black,
  fonttitle=\bfseries,
  boxrule=0.5pt,
  title={#1}
}

\newtcolorbox{thinkcodebox}[1][]{
  enhanced,
  breakable,
  colback=thinkbg,            
  colframe=black!25,
  colbacktitle=black!8,   
  boxrule=0.5pt,
  title={#1},
  fonttitle=\bfseries,
  coltitle=black
}

\lstdefinestyle{modelcode}{
  basicstyle=\ttfamily\small,
  columns=fullflexible,
  breaklines=true,
  backgroundcolor=\color{black!5},  
  frame=none
}

\newtcolorbox{sandboxbox}[1][]{
  enhanced,
  breakable,
  colback=darkyellow,
  colframe=black!25,
  colbacktitle=black!8,   
  boxrule=0.5pt,
  title={#1},
  fonttitle=\bfseries,
  coltitle=black
}

\newtcolorbox{answerbox}[1][]{
  enhanced,
  breakable,
  colback=darkblue,            
  colframe=black!25,
  colbacktitle=black!8,   
  boxrule=0.5pt,
  title={#1},
  fonttitle=\bfseries,
  coltitle=black
}

\definecolor{cvprblue}{rgb}{0.21,0.49,0.74}
\usepackage[pagebackref,breaklinks,colorlinks,allcolors=cvprblue]{hyperref}
\usepackage{float}
\def\confName{CVPR}
\def\confYear{2026}

\newcommand{\point}[1]{\noindent\textbf{#1.}}

\title{\model: Code with Images for Faithful Visual Reasoning\\via Tool-Aware Policy Optimization\vspace{-0.5em}}

\author{%
  Xinhai Hou\tsc{1*}\quad
  Shaoyuan Xu\tsc{2*}\quad
  Manan Biyani\tsc{2}\quad
  Moyan Li\tsc{2}\quad
  Jia Liu \tsc{3}\\
  Todd C. Hollon \tsc{1}\quad
  Bryan Wang\tsc{2$\dagger$}\\[0.25em]
  \small
  \tsc{1}University of Michigan\quad
  \tsc{2}Amazon.com\quad
  \tsc{3}The Ohio State University\quad\\
  \small
  \tsc{$\dagger$}Corresponding Author\quad\tsc{*}Equal contribution\\
  \small
    \faGithub\ \url{https://github.com/RenlyH/CodeV}\quad
}

\begin{document}
\maketitle


\vspace{-6pt}
\begin{abstract}
Agentic vision–language models are increasingly trained to “think with images” by calling image operations. However, we show that high final-answer accuracy often hides unfaithful visual reasoning: models may invoke tools on irrelevant regions or ignore tool outputs entirely, yet still guess the correct answer. In this work, we first propose a faithfulness evaluation protocol that measures whether intermediate visual tool outputs (e.g., crops) actually contain the queried evidence. This reveals that recent visual agents achieve high final-answer accuracy but exhibit low rates of faithful tool-use on visual search benchmarks. 
We then introduce CodeV, a code-based visual agent trained with Tool-Aware Policy Optimization (TAPO). 
TAPO is a process-level RL framework that augments GRPO with dense rewards defined directly on visual tool inputs and outputs, rather than on chain-of-thought tokens, making supervision easier to verify and less susceptible to reward hacking. 
CodeV represents visual tools as executable Python code, and TAPO assigns step-wise rewards based solely on the question and tool output, encouraging both necessary and evidence-consistent tool use. In a two-stage SFT+RL pipeline, CodeV achieves competitive or superior accuracy while substantially increasing faithful tool-use rates on related visual search benchmarks. Beyond visual search, CodeV attains strong performance on a range of multimodal reasoning and math benchmarks, suggesting that explicitly supervising intermediate tool behavior is crucial for building trustworthy, agentic visual reasoning systems.
\end{abstract}    
\vspace{-1em}
\section{Introduction}
\vspace{-3pt}
\begin{figure}[ht!]
    \centering
        
    \includegraphics[width=\columnwidth]{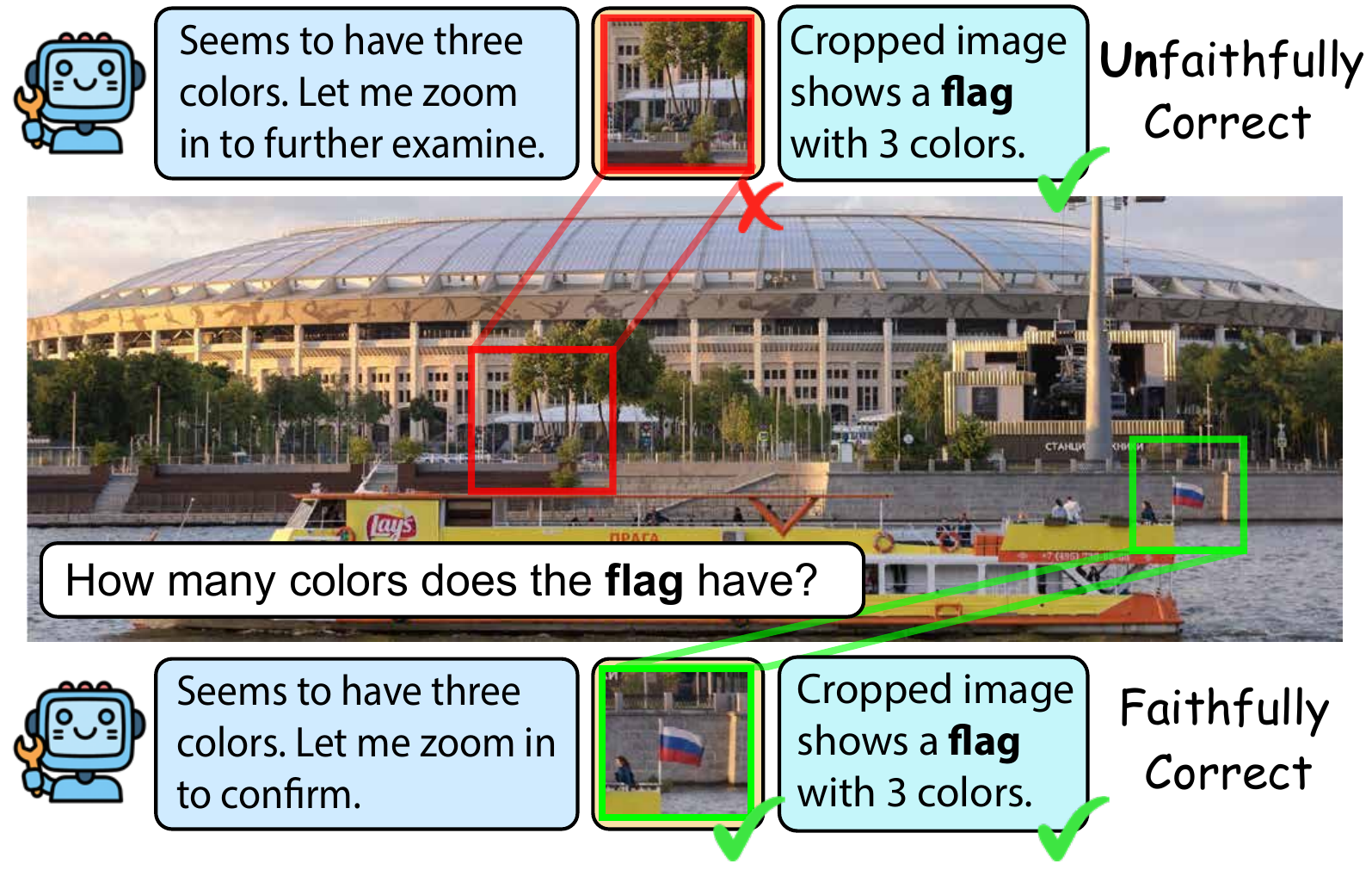}
    \vspace{-1.5em}
    \caption{An example of visual agentic system generating ``unfaithful'' trajectory: the cropping tool is used at the wrong region with unfaithful analysis but leads to correct answer.} 
    \label{fig:intro_unfaith}
    \vspace{-.2in}
\end{figure}

Large language models (LLMs) and vision–language models (VLMs) now tackle challenging tasks that require extended thinking and tool–integrated reasoning: web search, code execution, and agentic planning for complex goals \citep{jaech2024openai,openai2025thinkimage,guo2025deepseek,team2025kimi15,team2025kimik2,lao2025gptuner,zhang2024reactable,yao2022react,lao2024demonstration,yu2025correct,bamba2025xrpo,kondepudi2025health,lyu2026learning,zhao2025towards}. In the visual domain, current visual systems explicitly ``think with images''~\citep{openai2025thinkimage}, interleaving reasoning with visual actions such as cropping, segmentation, and other image operations to ground answers in tool responses \cite{gupta2023visual,suris2023vipergpt,hu2024visualsketchpad}, yielding notable gains on compositional tasks such as visual reasoning \citep{lu2023mathvista}, visual search \cite{wu2024v}, and chart reasoning \citep{wang2024charxiv}. This agentic reasoning paradigm consistently outperforms single-shot prediction and offers a scalable path to reliable, evidence-based solutions. However, many of the most capable agentic systems are based on proprietary models (GPT-4o, o3). In this work, we therefore focus on training open-source VLMs to acquire similar agentic visual tool-use behavior.

Reinforcement learning with verifiable rewards (RLVR) has emerged as a central approach for eliciting complex reasoning \citep{jaech2024openai,guo2025deepseek,team2025kimi15}. 
Recent work extends RLVR to agentic visual reasoning, where agents learn to interleave image operations (e.g., cropping, segmentation, depth estimation) with textual reasoning \citep{su2025pixelreasoner,zheng2025deepeyes,su2025openthinkimg,huang2025visualtoolagent}. Exposing images through a code interpreter further allows the model to invoke a rich library of image operations (e.g., crop, rotate) and leverage its visual capability over multiple tool calls, as demonstrated by OpenAI o3 \citep{openai2025thinkimage} and related works \citep{zhang2025thyme}. This code-based interface avoids reliance on external, heavyweight tool API and leverages the code patterns the model has already seen at scale during pretraining, making tool use both more expressive and more natural. 
Despite these advances, we observe a key failure mode in current open-source visual agents. Models such as DeepEyes~\citep{zheng2025deepeyes} and Pixel-Reasoner~\citep{su2025pixelreasoner} can score highly on visual search benchmarks while relying on incorrect or misaligned tool calls (e.g., cropping the wrong region but still answering correctly; Figure~\ref{fig:intro_unfaith}). This motivates a faithfulness analysis in Section~\ref{sec:tool_faithful}, where we systematically evaluate \emph{visual tool use} by checking whether intermediate image operations (e.g., crops) actually contain the object or region referenced in the question, thereby measuring action-level faithfulness for visual search.

Given these experimental results, we hypothesize that the observed misalignment arises from reward hacking~\citep{10.5555/3600270.3600957}, driven by two key issues in reward design: (1) outcome dominance and (2) reward sparsity. Specifically, current rewards are \emph{outcome-dominant}: they prioritize final accuracy or whether a tool was invoked, while providing no supervision on how the tool was used~\citep{zheng2025deepeyes,feng2025retool}. Lacking step-level credit assignment, policies quickly exploit this reward signal by hallucinating tool use or performing meaningless operations \cite{valentin2024cost}. This mirrors the patterns in web-based agents that issue superficial search queries or clicks while largely ignoring retrieved evidence \cite{yao2022react,li2025search,nakano2021webgpt,zhang2024reactable}. Second, the reward is \emph{sparse}, producing unstable optimization dynamics. Early-stage rollouts on difficult tasks frequently receive zero rewards, which discourages the faithful use of tools with partially correct attempts. Although reward shaping or invocation bonuses push the policy to use tools, they also introduce reward hacking, encouraging unfaithful behaviors such as unnecessary image operations and trivial computations ~\citep{liu2025visualarft,zheng2025deepeyes,zhou2025reinforced}. As models scale up and reinforcement learning (RL) training intensifies, these issues become increasingly pronounced, amplifying the tendency for models to optimize for superficial success signals rather than genuine reasoning faithfulness. These observations of unfaithful tool use and unstable training motivate the central research question of this paper: 
\vspace{-2pt}

\begin{tcolorbox}[left=1.2pt,right=1.2pt,top=1.2pt,bottom=1.2pt]
How do we incentivize vision–language models to produce faithful reasoning grounded in tool outputs?
\end{tcolorbox}

\vspace{-2pt}
To answer this question and elicit more powerful coding capability, we rethink RLVR from a \emph{process reward} perspective and propose \underline{T}ool-\underline{A}ware \underline{P}olicy \underline{O}ptimization (\algo), which treats tool use as a sequence of verifiable decisions. Rather than rewarding only the final answer, \algo leverages a rule-based check and a judge model to assign \emph{step-level rewards} to each \textit{tool output} that requires context of question and answers to align. \algo couples the scalability of GRPO with dense, process-aware signals and avoids reliance on training a separate, hack-prone reward model to monitor unverifiable chain of thought. 
The TAPO-based reinforcement learning (RL) is further integrated with a cold-start instruction tuning phase to form a two-stage curriculum training framework,
based on which wWe adopt and further curate the commonly used instruction tuning and RL datasets \citep{zheng2025deepeyes,wang2025sota,zhang2025thyme}. Trained on these datasets, our model \model-7B, built on top of Qwen2.5-VL-7B \citep{bai2025qwen2}, yields stable, faithful tool use, reduces tool-related failure cascades, and achieves strong performance across 10 challenging benchmarks, narrowing the gap to proprietary systems \citep{openai2025thinkimage,jaech2024openai}. The key contributions of our work are summarized as follows:

\begin{enumerate}
  \item \textbf{Tool-Aware Policy Optimization (\algo).} We introduce \algo, a novel policy optimization method that augments GRPO with dense process rewards for tool use necessity and tool output consistency. \algo delivers markedly better sample efficiency and training stability.
  \item \textbf{Faithfulness evaluation for visual agents.} We propose a protocol that detects unfaithful visual reasoning, yielding quantitative faithfulness metrics beyond accuracy.
  \item \textbf{Empirical gains in faithfulness and accuracy.} Training \model with \algo largely improves tool faithfulness for the visual reasoning task and demonstrates strong performance on 10 challenging benchmarks.
  \item \textbf{Open recipe and resources.} We release the full training recipe for \model, including the implementations of \algo, along with datasets and the corresponding model to facilitate reproduction and extension.
\end{enumerate}
\vspace{-2pt}
\section{Preliminaries}
\vspace{-3pt}
\subsection{Reinforcement Llearning for Rreasoning LLMs}
\label{sec:prelim_rl4llm}
RLVR ~\citep{shao2024deepseekmath,lightman2023let} has proven effective for eliciting  multi-step reasoning in LLMs \citep{guo2025deepseek, jaech2024openai}. 
In this paradigm, the model generates its own trajectories, receives feedback from the reward signal, and iteratively refines its policy. Formally, given a dataset $\mathcal{D}$ of prompts $q$, the policy $\pi_\theta$ generates an output $o$ and receives a reward $R(q,o)$. We optimize:
\vspace{-3pt}
\begin{align}
\max_{\pi_\theta} \;
& \mathbb{E}_{x \sim \mathcal{D}, \, o \sim \pi_\theta(\cdot|q)}
 R(q, o) \nonumber \\
& - \beta \, \mathbb{D}_{\mathrm{KL}}\big(
   \pi_\theta(o|q) \, \| \, \pi_{\mathrm{ref}}(o|q)
 \big),
\label{eq:rl-objective}
\end{align}
\vspace{-3pt}
where $\mathbb{D}_\mathrm{KL}$ serves as a regularization penalty that constrains the updated policy to stay close to the reference policy, thereby stabilizing training in policy-gradient methods such as PPO and GRPO. In standard outcome-only setups, $R$ is used at the final answer, providing no signal on which intermediate tool decisions mattered, which becomes spurious in agentic RL, particularly when visual actions are required. The KL term keeps $\pi_\theta$ close to the reference policy, which stabilizes training and preserves useful SFT priors such as sensible tool-usage patterns.

\subsection{Agentic Visual Reasoning }\label{sec:prelim_avr}
We consider the problem of answering visual questions through iterative reasoning with tool use. Given an input $\mathrm{x}=(\mathit{V,Q})$ comprising an image $\mathit{V}$ and a textual query $\mathit{Q}$, a policy model $\mathrm{\pi_\theta}$ generates a trajectory for at most $T$ turns of conversation:
$\tau = (\mathbf{x}, a_1, o_1, a_2, o_2, \ldots, a_T)$
where each action \(a_t \sim \pi_\theta(\ \cdot \mid \mathbf{x}, h_{t-1})\) 
conditions on the input and history 
\(h_{t-1} = \{(a_i, o_i)\}_{i=1}^{t-1}\).
Actions $a_t$ at $t$ turn belong to one of the three categories:
    {\texttt{<think>}} free-form textual thoughts.
{\texttt{<tool\_call>}} executable token that invokes a tool.
{\texttt{<answer>}} final answer that ends the trajectory.
The observation $o_t$ is available only when $a_t$ is a \texttt{<tool\_call>} that produces an output (e.g., transformed images, numbers, or errors); for \texttt{<think>} and \texttt{<answer>} actions, we set $o_t$ to null.
This unified action space enables the flexible interleaving of reasoning and tool use, allowing the model to dynamically adjust its strategy based on intermediate results rather than follow a fixed workflow. Our formulation treats agentic visual reasoning as a single, cohesive decision process, where each action is conditioned on the full trajectory history, including all previous execution outcomes.

Most recent works follow a two-stage curriculum pipeline \cite{openai2025thinkimage,huang2025visualtoolagent,liu2025visualarft,zhang2025thyme,su2025pixelreasoner,feng2025retool,chu2025sft,li2025torl}: supervised fine-tuning (SFT) and then a reinforcement learning stage. SFT exposes the model to high-quality traces in which tools are invoked, and their outputs are woven into the reasoning, teaching basic patterns of tool use. Many papers highlight the importance of high-quality and diverse trajectory data, such as trial-and-error with self-reflection \cite{su2025pixelreasoner}. 
Then, RL is used to optimize an explicit reward, typically based on final-answer correctness, along with a tool invocation bonus, encouraging exploration under distribution shift while consolidating high-reward behaviors on the target data distribution. We acknowledge the importance of SFT as a cold-start step. However, without proper handling at the RL stage, it can easily diverge and result in unfaithful trajectories.
\begin{figure}[ht]
    \centering
    \includegraphics[width=0.8\columnwidth]{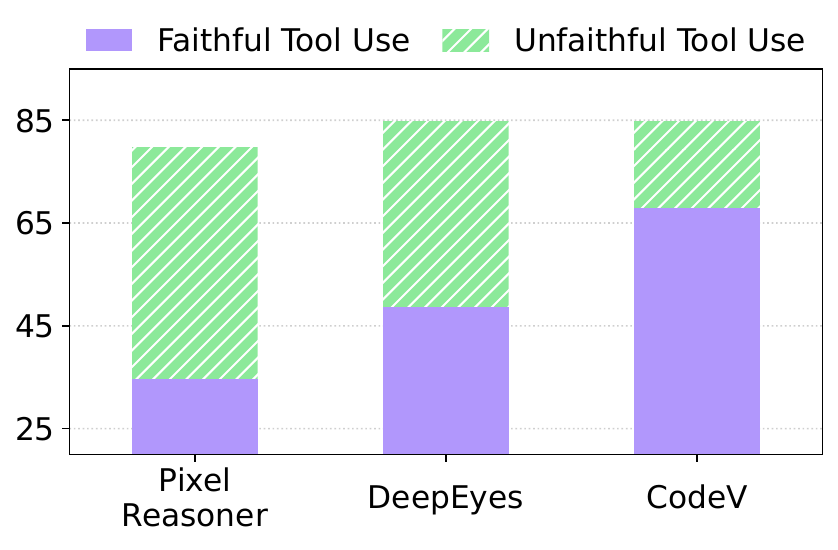}
    \vspace{-0.5em}
    \caption{\textbf{Faithfulness conditioned on correct answers in V* \cite{wu2024v} Benchmark.} For this visual search problem, crop is treated as the most effective tool use \cite{zhao2025pyvision,zheng2025deepeyes,su2025pixelreasoner}. Therefore, we define faithful tool use as cropped image from tool use capturing \textit{any} target object and evaluate how many correct answers are also faithful, as shown in \textcolor[HTML]{B197FC}{violet}. For the remaining correct answers, the tool uses do not capture the target object and are treated as unfaithful, as shown in \textcolor[HTML]{8ce99a}{mint}. Recent visual agents achieve high final-answer accuracy but fail to use tools faithfully. \model shows great improvement in faithfulness with no decrease in accuracy.}
    \vspace{-1em}
    \label{fig:faith_analysis_vstar}
\end{figure}

\section{Evidence of Unfaithful Tool Use in VLMs}\label{sec:tool_faithful}

Faithful reasoning is central to building trustworthy agentic AI systems. When a model performs agentic visual reasoning and invokes visual tools, its intermediate actions should accurately reflect and depend on the information returned by those tools. As shown in Figure~\ref{fig:intro_unfaith}, without such alignment, a model can produce correct answers for the wrong reasons, undermining both interpretability and reliability. In the visual setting, this lack of faithfulness makes it unclear whether the model genuinely ``sees'' the evidence or is instead exploiting text cues to produce plausible outputs.

To study this behavior, we design experiments that quantitatively evaluate how faithfully models ground their reasoning in \emph{visual tool use}. Rather than inspecting model-generated chain-of-thought tokens, we focus on the consequences of tool actions, which are easier to verify. Specifically, we test whether intermediate tool outputs (e.g., cropped images) remain relevant to the question, providing a direct proxy for whether the model’s reasoning follows the intended problem-solving path.

\smallskip
\point{1) Baselines and benchmarks}
We evaluate two state-of-the-art recently open-sourced models, DeepEyes~\cite{zheng2025deepeyes} and Pixel-Reasoner~\cite{su2025pixelreasoner}, which are trained for agentic visual reasoning through cropping-based tool use. Experiments are conducted on visual search benchmarks where a genuine answer should localize a small object in a high-resolution image and then describe its attributes or relations. We primarily use V*~\cite{wu2024v}, and additionally report results on HRBench-4K~\cite{wang2025hrbench} and Thyme~\cite{zhang2025thyme} in Section~\ref{sec:exp:eval_faithfulness}.

\smallskip
\point{2) Evaluation setup}
To diagnose the faithfulness of visual tool use in these models, we evaluate whether intermediate cropped images actually contain the object or region referenced in the question. This directly targets their cropping-based design, where image operations are the core mechanism for grounding reasoning in visual evidence~\cite{zhao2025pyvision}. Concretely, for each example, we provide the original question together with an \emph{intermediate} tool-output crop to GPT-4o, which we use as a judge, and prompt it to decide whether the crop includes the queried object or region. When a model produces multiple crops, we regard the visual tool use as faithful if \textit{any} crop is judged to contain the target. We then report the proportion of faithful versus unfaithful tool calls, conditioned on the final answer being correct. A higher faithful tool-use rate reflects a more reliable, action-level alignment between the model’s reasoning, the intended visual evidence, and the task objective. Detailed evaluation settings, including prompt construction, are provided in the Appendix~\ref{supp:faith}.

\smallskip
\point{3) Results}
Figure~\ref{fig:faith_analysis_vstar} shows the proportions of \emph{faithful} and \emph{unfaithful} tool calls under the condition that the final answer is correct. While they have a relatively high score on final answer accuracy, only about half of the intermediate tool calls are judged to be faithful to the original question (57\% for DeepEyes and 43\% for Pixel-Reasoner).

\smallskip
\point{4) Takeaways}
These results reveal that high final-answer accuracy does \emph{not} guarantee strong agentic visual reasoning: models can often guess the correct answer from text cues like answer options, while using visual tools in unfaithful ways. In other words, current benchmarks can substantially overestimate the underlying visual tool-use capability. This suggests that the training objective should not only ensure final correctness but also enforce the correctness of tool use, both when constructing SFT dataset and when designing RL rewards. Aligning intermediate tool use with the intended evidence is, therefore, crucial for developing genuinely agentic and trustworthy visual reasoning systems.

\section{Methodology}
\begin{figure*}[ht!]
    \centering
    \includegraphics[width=\textwidth]{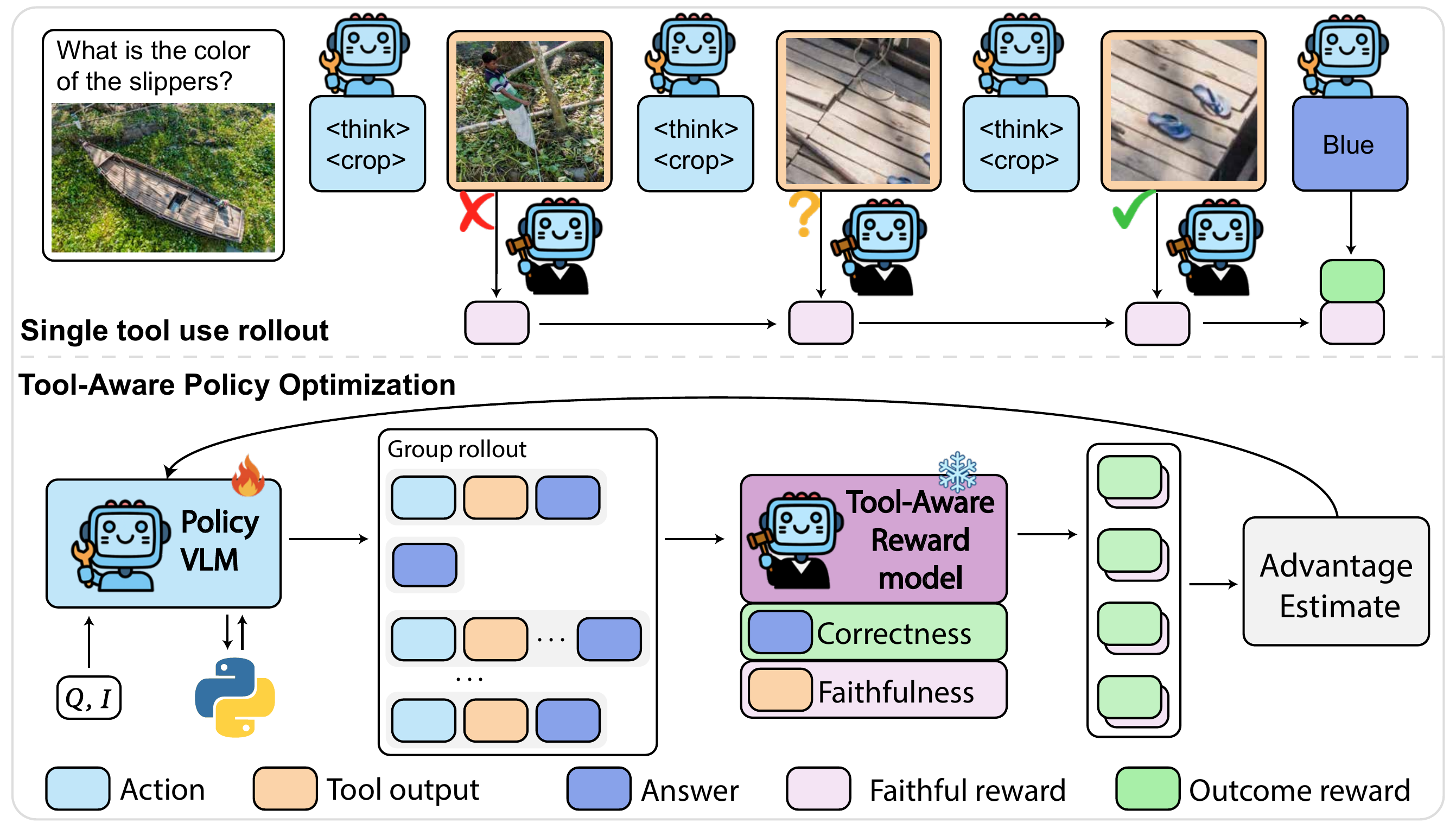}
    \vspace{-1.5em}
    \caption{Overview of the CodeV rollout and Tool-Aware Policy Optimization (TAPO). The model processes an image $I$ and question \textit{Q} pair, using tools like cropping to generate intermediate results for its final answer. Tool faithfulness will be scored by a reward model. For the tool like cropping, reward model will score $r^\mathrm{tool}$ based on the observability of the target object in the cropped image. The final answer correctness will be used as outcome reward. The policy VLM is fine-tuned with tool-aware policy optimization, a GRPO-style reinforcement learning approach. The policy VLM will conduct multiple rollouts for the same \textit{Q} and $I$ with tool use. These rollouts will be scored by the hybrid reward system that combines faithfulness and correctness. Final reward will be normalized within the group and used to estimate relative advantage for the policy VLM to update.}
    \vspace{-0.85em}
    \label{fig:overview_codev}
\end{figure*}
In this section, we instantiate the general agentic visual reasoning formulation in Section~\ref{sec:prelim_avr}  with an agent that generates \underline{Code} for \underline{V}isual reasoning (\model), and describe our \underline{T}ool-\underline{A}ware \underline{P}olicy \underline{O}ptimization (\algo) framework.

\subsection{Overview}\label{sec:overview}
Figure~\ref{fig:overview_codev} summarizes the training pipeline of \model. It consists of two main stages: supervised fine-tuning (SFT) and reinforcement learning (RL) with Tool-Aware Policy Optimization (TAPO). The goal is to optimize the model for both faithfulness and accuracy in visual reasoning tasks.

In the first stage, SFT initializes the model by training it on datasets involving image operations (e.g., cropping, rotating) and multi-round refinement tasks. This stage ensures efficient tool use and minimizes reliance on direct answers, with the learned knowledge serving as the reference policy.

In the second stage, similar to GRPO, TAPO performs on-policy rollouts and rewards are computed using a Monte Carlo-based baseline. TAPO balances tool use and accuracy, guiding the model to generate faithful tool steps while avoiding unnecessary or unfaithful actions.

The reward system combines accuracy and faithfulness, rewarding useful tool outputs (e.g., relevant crops) and penalizing ``lazy'' tool use, such as large, uninformative crops or incorrect tool operations.

\subsection{Code-based agentic rollouts}
\label{sec:method_rollouts}
As in Section~\ref{sec:prelim_avr} and Figure \ref{fig:overview_codev}, a rollout is
\begin{equation}
\tau = (\mathbf{x}, a_1, o_1, \ldots, a_T), \quad 
\mathbf{x} = (V, Q),
\end{equation}
where $a_t \sim \pi_\theta(\cdot \mid \mathbf{x}, h_{t-1})$ and 
$h_{t-1} = \{(a_i,o_i)\}_{i=1}^{t-1}$.
In \model, each $a_t$ is one of:
\begin{itemize}
    \item \texttt{<think>}: free-form text for intermediate reasoning;
    \item \texttt{<code>}: a delimited Python program that operates on $V$;
    \item \texttt{<answer>}: final answer that terminates $\tau$.
\end{itemize}
Only completed \texttt{<code>} blocks are executed in a restricted Python sandbox with read-only access to $V$ and a small set of deterministic image / math utilities.
Execution returns an observation $o_t$ (logs and optional derived images) that is appended to the context; for \texttt{<think>} and \texttt{<answer>} we set $o_t=\varnothing$.
Thus, tool use is fully represented inside the token-level policy over this mixed action space, without an external controller.
Implementation details of the sandbox are deferred to Appendix~\ref{sec:supp:sandbox}.

\subsection{TAPO: Tool-Aware Policy Optimization}
\label{sec:method_tapo}
We adopt a two-stage SFT+RL pipeline to train \model.

\point{Stage 1: SFT}
We highlight the importance of the cold-start SFT stage. We find that popular instruct-tuned models such as Qwen2.5-VL cannot transform the knowledge of coding into problem-solving capabilities: when performing RL with these models, rollouts using code on average have a lower correct rate than direct answering. The model will quickly converge to pure text reasoning and fail to conduct agentic visual reasoning. To elicit the knowledge of ``Code with Image'', the model needs to be cold-started via SFT. The SFT model serves as both initialization and as the reference policy $\pi_{\mathrm{ref}}$ in Eq.~\eqref{eq:rl-objective}.

\point{Stage 2: GRPO with group baselines}
We then perform GRPO-style optimization on on-policy rollouts.
For each prompt group $g$ (corresponding to a single image--question pair), we sample $K$ trajectories$\{\tau_k^{(g)}\}_{k=1}^K \sim \pi_\theta(\cdot \mid g),$
compute their rewards $R(\tau_k^{(g)})$, and define a \emph{group baseline} as a simple Monte Carlo critic as $b(g) = \frac{1}{K} \sum_{k=1}^K R\big(\tau_k^{(g)}\big).$
Given a trajectory $\tau$, let $g(\tau)$ denote its associated prompt group.
We define the trajectory-level advantage as $
A_t = R(\tau) - b\big(g(\tau)\big),
$
and broadcast this value to all time steps along the trajectory for every time step $t$ in $\tau$.

For every token-level action $a_t$ in state $s_t$ (the full multimodal history up to time $t$), we form the importance ratio
\begin{equation}
r_t(\theta)
= \frac{\pi_\theta(a_t \mid s_t)}{\pi_{\mathrm{ref}}(a_t \mid s_t)}.
\end{equation}
TAPO optimizes a clipped GRPO objective
\begin{align}
\mathbb{E}_{\tau,t}
\Big[
&\min\big(
r_t(\theta) \, A_t,\;
\mathrm{clip}(r_t(\theta), 1-\epsilon, 1+\epsilon) \, A_t
\big)
\Big]
\nonumber\\
&\quad
- \beta \,
\mathbb{E}_{q}
\Big[
\mathbb{D}_{\mathrm{KL}}\big(
\pi_\theta(\cdot \mid q)
\big\|
\pi_{\mathrm{ref}}(\cdot \mid q)
\big)
\Big],
\end{align}
which instantiates Eq.~\eqref{eq:rl-objective} with group-wise advantages and PPO-style clipping applied to trajectories that contain both language and code actions. Under high-temperature sampling, group sampling implicitly attributes credit: since trajectories differ at certain steps and better trajectories yield higher $R(\tau)$, the resulting advantage $A(\tau)$ will selectively increase the probability of those decisive tool-use steps.

\paragraph{Reward design for TAPO.}
\label{sec:method_reward}
TAPO employs a hybrid reward that combines final-answer correctness with the faithfulness of visual tool use:
\begin{equation}
R(\tau)
= \lambda_{\mathrm{acc}} \, r^{\mathrm{acc}}(\tau)
+ \lambda_{\mathrm{tool}} \, r^{\mathrm{tool}}(\tau),
\end{equation}
where $r^{\mathrm{acc}}$ measures answer quality and $r^{\mathrm{tool}}$ evaluates tool steps, and $\lambda_{\mathrm{acc}}, \lambda_{\mathrm{tool}} \in [0,1]$ are weighting coefficients.
Let $\mathcal{T}_{\mathrm{tool}}$ be the set of indices where $a_t$ is a \texttt{<code>} action and aggregate
\begin{equation}
r^{\mathrm{tool}}(\tau)
= \frac{1}{|\mathcal{T}_{\mathrm{tool}}|}
  \sum_{t \in \mathcal{T}_{\mathrm{tool}}}
  r^{\mathrm{tool}}_t.
\end{equation}

\paragraph{Answer reward.}
For $r^{\mathrm{acc}}$, we use exact or programmatic matching when ground-truth answers are available (e.g., numeric or categorical VQA), and an LLM-as-a-judge for open-ended tasks. In open-ended visual reasoning, final correctness already relies on an external judge \cite{zheng2025deepeyes,zhang2025thyme}, so TAPO reuses the same infrastructure without extra deployment burden.

\paragraph{Tool reward as evidence checking.}
Early experiments showed that directly judging the \emph{model's} code and thoughts is both brittle and expensive.
Instead, TAPO only inspects \emph{non-model-generated} context:
$
(Q,\; o_t,\; \text{metadata}),
$
where $Q$ is the original question and $o_t$ is the sandbox output (cropped image or its description, coordinates, simple statistics).
The judge template is asked whether the current tool result provides \emph{any} useful evidence for answering $Q$.
In particular, the rubric:
(i) encourages any crop that clearly contains a relevant object or region; and
(ii) does not require the crop to include all relevant objects.
For questions involving multiple targets, a trajectory that isolates at least one correct region is rewarded rather than penalized.

This evidence-based view is easier to verify than diagnosing internal reasoning:
the judge never sees the chain-of-thought or code and only answers ``does this piece of evidence help with this question?''.
IoU-based rewards against ground-truth boxes would be an alternative, but they require dense annotations and do not generalize to tasks without bounding boxes.
Our judge-based formulation works on top of arbitrary tool outputs and can be reused for other task families with minimal changes. As for the choice of judge model, we leverage Qwen2.5-VL-32B as our judge with ablation on GPT-5-nano with results shown in \ref{exp:ablation}.

\paragraph{Redlines and discouraging lazy crops.}
To avoid reward hacking and trivial large crops, we combine 2 mechanisms:
\begin{itemize}
    \item \textbf{SFT shaping.} SFT-stage already teaches localized, high-resolution crops and multi-round refinement: large, uninformative boxes are rare in the training traces.
    RL fine-tunes around this prior rather than discovering cropping behavior from scratch.
    \item \textbf{Judge rubric.} For each crop, the judge is instructed to focus on whether the \emph{main content} of the cropped image clearly contains the object or region needed to answer $Q$.
    Extremely large or cluttered crops, where the target is small or barely visible, typically fail this check and receive low $r^{\mathrm{tool}}_t$.
    We also treat obvious misuse of tools (e.g., using image read/save as scratchpad) as red flags and assign negative reward.
\end{itemize}
Collectively, these techniques discourage lazy crops.
At the same time, we keep $|\lambda_{\mathrm{tool}}| < |\lambda_{\mathrm{acc}}|$, so global answer correctness remains the dominant training signal.

\paragraph{Tasks without explicit visual search.}
For tasks that do not require localized visual search (e.g., some counting or captioning problems), dense tool rewards can easily encourage unnecessary tool calls and reward hacking. If no tool is needed to answer correctly, the optimal policy is simply to skip tools. Here, we deliberately keep $r^{\mathrm{tool}}_t$ close to zero by default and only apply a negative reward for clear redlines such as invalid coordinates, repeated no-op crops, or avoidable sandbox errors. 
\section{Experiments}
We present our training data and implementation details in \ref{sec:exp:setup} and provide an overview of the selected baselines and evaluation benchmarks in Section \ref{exp:main}. We conduct additional faithfulness evaluation in Section \ref{sec:exp:eval_faithfulness} and finally we provide an ablation study in \ref{exp:ablation}. Additional qualitative examples will be presented in Appendix \ref{sec:supp:qualitative}.

\begin{figure*}[ht!]
    \centering
    \includegraphics[width=\textwidth]{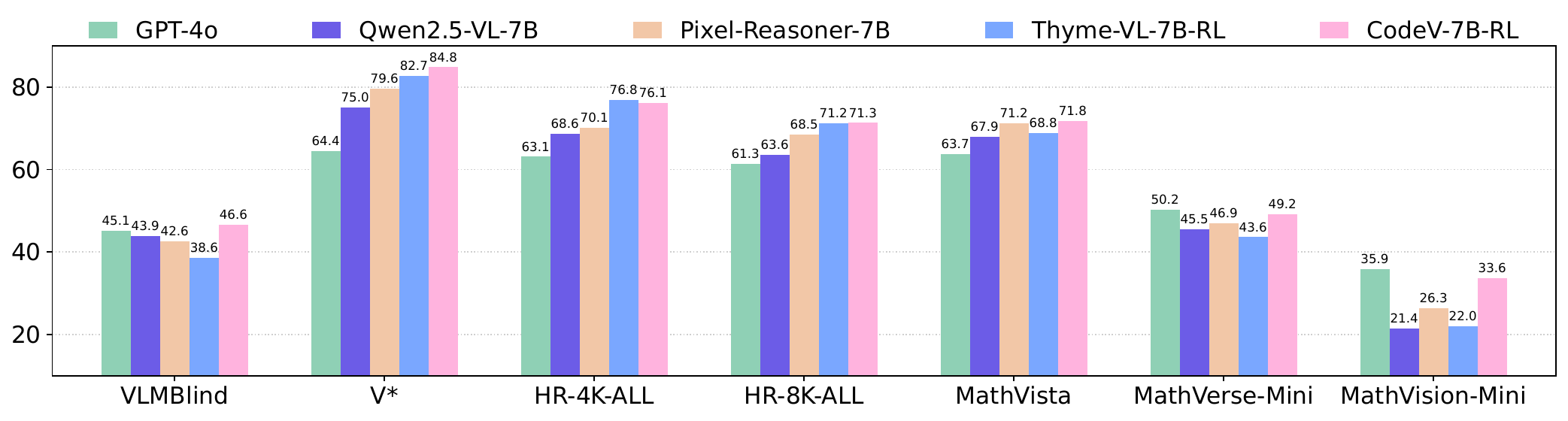}
    \vspace{-2em}
    \caption{Performance across primitive perception, visual search, reasoning and math benchmarks. All model other than GPT-4o are 7B model with proper setup of tool use.}
    \vspace{-1.em}
    \label{fig:exp_1}
\end{figure*}

\subsection{Experimental Setup}\label{exp:data}\label{sec:exp:setup}\label{method:data_prep}

\point{Data preparation} High-quality training data is crucial for incentivizing visual reasoning and enabling the model to solve complex tasks. We use Thyme-SFT \citep{zhang2025thyme}, a curated dataset with rich multi-step reasoning demonstrations in code and direct-answer samples. For RL training, we construct a larger dataset from open-source sources such as Thinklite-70K \citep{wang2025sota} and DeepEyes-47K \citep{zheng2025deepeyes}, first removing problems requiring external knowledge (e.g., OK-VQA \citep{marino2019ok}) and then rigorously cleaning noisy labels via automated checks with Qwen2.5-VL-32B and manual verification. We also prompt Qwen2.5-VL-7B to answer each question eight times and discard samples whose empirical accuracy exceeds 0.9 during sampling, which improves RL efficiency since those rollouts receive the same reward and no relative advantage. This dataset is used exclusively for our TAPO-based RL stage. Details are in Appendix~\ref{sec:supp:data}.

\point{Implementation details}
For the SFT stage, we follow the setting in Thyme \cite{zhang2025thyme} and finetune for three epochs. As for the RL stage, we design our prompt template with the system prompt explicitly specifying the required format, including the special tokens \texttt{\textless think\textgreater}, \texttt{\textless code\textgreater}, \texttt{and \textless answer\textgreater} , along with the sandbox output token as \texttt{\textless sandbox\_output\textgreater}. Detailed prompt configurations are provided in Appendix \ref{sec:supp:prompt}. 
For training hyperparameters, both the rollout batch size and the training batch size are set to 256, with eight rollouts generated per sample and 200 steps of updates. Sampling is conducted at a temperature of 1.0, and optimization is performed with a learning rate of $1\times10^{-6}$. Our experiment uses  eight H200 GPUs, with 255 and 288 GPU hours in the SFT and RL stages, respectively.

\subsection{Performance of CodeV}\label{exp:main} 
We evaluate \model against both open-source baselines (Qwen2.5-VL-7B, Pixel-Reasoner-7B, Thyme-VL-7B-RL) and a proprietary model (GPT-4o) on a suite of comprehensive multimodal perception and reasoning benchmarks.

\smallskip
\point{Results} 
Figure~\ref{fig:exp_1} summarizes the results (full numbers in Appendix~\ref{sec:supp:full_result}, Tables~\ref{tab:perception_results} and~\ref{tab:math_reason_results}). On VLMBlinds, \model attains the best score (46.7), improving over Qwen2.5-VL-7B and Pixel-Reasoner-7B by roughly 3 points and over Thyme-7B-RL by about 8 points, and also slightly surpassing GPT-4o. This suggests that training with TAPO does not sacrifice performance on primitive but challenging perception tasks.
On large-image visual search benchmarks (V*, HRBench-4K-all, HRBench-8K-all), \model delivers strong overall performance. It achieves the highest score on V* (84.8), clearly outperforming GPT-4o (64.4) and Qwen2.5-VL-7B (75.0). On HRBench-4K-all and HRBench-8K-all, \model matches or slightly exceeds the best open-source baseline while maintaining a sizable margin over GPT-4o, indicating robust fine-grained visual reasoning at high resolution.
On math-heavy datasets, \model reaches 71.8 on MathVista, the best among all models, and 49.2 on MathVerse-Mini, closing most of the gap to GPT-4o while outperforming other open-source baselines. On MathVision-mini, \model substantially improves over Qwen2.5-VL-7B and Pixel-Reasoner-7B and remains competitive with GPT-4o.

Overall, these results show that \model narrows the gap between open-source and proprietary systems on both perception and reasoning tasks. The consistent gains over the strong Qwen2.5-VL-7B baseline suggest that incentivizing visual tool use with dense process-level rewards in \algo provides a principled path to more capable agentic VLMs.

\subsection{Evaluating Reasoning Faithfulness}\label{sec:exp:eval_faithfulness}
    \vspace{-1em}
\begin{figure}[h]
    \centering
    \begin{subfigure}[t]{\columnwidth}
        \centering
        \includegraphics[width=\linewidth]{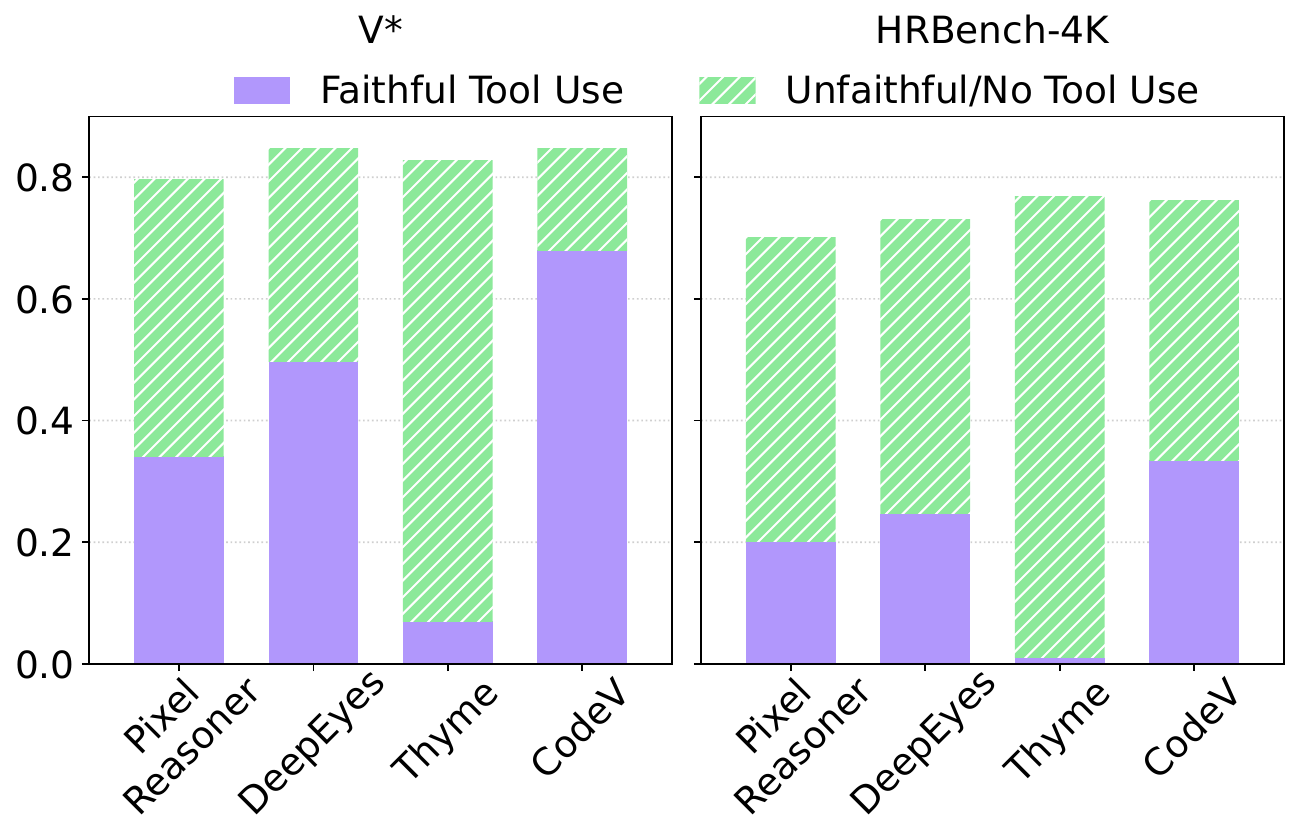}
        \label{fig:result:unfaith_combine}
    \end{subfigure}
    \vspace{-1em}
    \caption{Faithfulness comparison on V* and HRBench-4k benchmarks. The extremely low faithful tool use rate in \cite{zhang2025thyme} results from low tool use rate and decorative tool use in chain of thought.}
    \label{fig:result:unfaith_combined}
\end{figure}

To evaluate the faithfulness enabled by the design of TAPO, we extend the analysis originally presented in Section~\ref{sec:tool_faithful} to compare the faithfulness of baseline approaches against CodeV. Additionally, we add Thyme \cite{zhang2025thyme} and the HRBench-4K benchmark. We report the faithful rate alongside the unfaithful and no-tool-use rates. A higher rate of faithful tool-use indicates better alignment between the model’s reasoning and the question intent.

\smallskip
\point{Results} Figure~\ref{fig:result:unfaith_combined} compares faithful tool-use rates across the V* and HRBench-4k benchmarks. \model consistently attains the highest faithfulness, improving tool-use rates over Pixel-Reasoner and DeepEyes by double-digit margins on most settings and by more than 30 points in the most extreme case, while maintaining comparable or better answer accuracy. In relative terms, CodeV’s faithful tool-use rate is roughly 1.3--2$\times$ that of other tool-using baselines, indicating that TAPO effectively encourages the model to ground its decisions in the evidence returned by tools.

Thyme, in contrast, exhibits extremely low faithfulness on both datasets (single-digit points). Manual inspection shows that fewer than 10\% of its rollouts actually call visual tools: it often relies on purely textual reasoning, implicitly assuming a ``cropped image'' without executing the crop. These findings underscore the importance of explicit faithfulness evaluation and process-aware RL: without step-level incentives, models tend to optimize final accuracy while shortcutting tool use. By jointly rewarding answer correctness and faithful tool calls, \algo enables \model to sustain high accuracy while substantially improving the reliability of its visual reasoning. Additional tool perturbation study and tool use analysis are in Appendix \ref{sec:supp:perturb} and \ref{sec:supp:tool_stats}.

\subsection{Ablation Studies}\label{exp:ablation}
To better understand the contributions of our design choices, we conduct ablation studies on two critical aspects: (1) training stage ablation and (2) reward design in TAPO. We report the average scores across all perception and reasoning benchmarks. The full results and training curves are available in Appendix~\ref{sec:supp:ablation} and \ref{sec:supp:training_dynamics}. 

\smallskip
\point{Training stage ablation}
Starting from Qwen2.5-VL-7B, Zero-RL (without cold-start SFT stage) improves average performance by roughly 3--4 points but quickly collapses to pure text reasoning with minimal tool use, similar to the learning trap observed in Pixel-Reasoner~\cite{su2025pixelreasoner}. The cold-start SFT model slightly underperforms these baselines but produces trajectories with richer tool use. Building on this initialization, \model trained with TAPO further improves average scores by about 1--3 points over Zero-RL and 6--8 points over SFT, with the largest gains on perception benchmarks, indicating that code and sandbox feedback are particularly beneficial for visual reasoning.

\smallskip
\begin{table}[ht]
\centering
\begin{tabular}{lccc}
\toprule
Training stage    & \textbf{Reasoning} & \textbf{Perception} \\ \hline
Qwen2.5-VL-7B     & 49.7  & 62.8 \\
Zero-RL           & 52.9 &  67.0 \\
Cold-start SFT    & 47.7  & 61.7 \\
\model            & 54.2  & 69.7 \\ 
\toprule
\end{tabular}
\caption{\textbf{Training stage ablation.} Average performance across all reasoning and perception benchmarks.}\label{sec:tab:train_stage}    
\end{table}

\smallskip
\begin{table}[ht]
\centering
\begin{tabular}{lccc}
\toprule

Reward design     & \textbf{Reasoning} & \textbf{Perception} \\ \hline

\ \ \  Accuracy only & 51.2 & 67.5 \\
+ Consistency      & 50.4   & 67.6 \\
+ GPT-5-nano Judge & 52.4   & 68.7 \\
\model             & 54.2   & 69.7 \\ 
\toprule
\vspace{-1em}
\end{tabular}
\caption{\textbf{Reward design ablation.} Average performance across all reasoning and perception benchmarks.}\label{sec:tab:reward_design}
\vspace{-0.5em}
\end{table}

\begin{figure}[h]
    \centering
    \begin{subfigure}[t]{\columnwidth}
        \centering
        \includegraphics[width=0.8\linewidth]{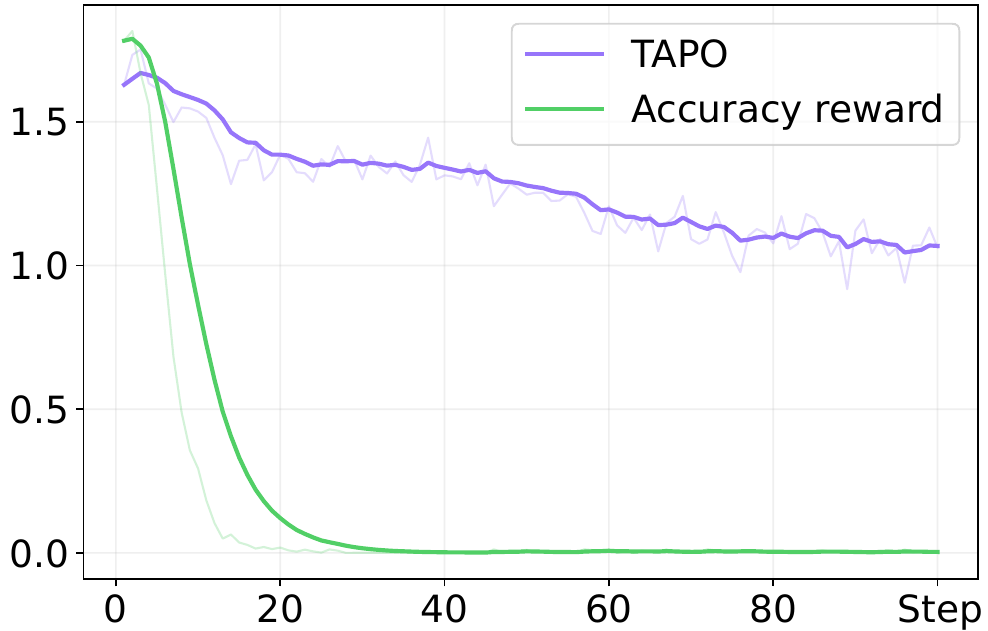}
    \end{subfigure}
    \vspace{-1.5em}
    \caption{Tool invocation count during early RL training.}\label{fig:result:rl}
\end{figure}

Table~\ref{sec:tab:reward_design} studies reward design starting from the cold-start SFT model. An accuracy-only outcome reward yields a small gain over SFT, but eventually drives the policy toward text-only reasoning (green curve in Figure~\ref{fig:result:rl}), and adding a consistency reward, similar as \cite{zhang2025thyme}, only has negligible changes in average performance. Using GPT-5-nano as a judge within TAPO provides an improvement of approximately 2 points in reasoning and 1 point in perception. The full TAPO reward used in \model adds another \(\sim\)2-point gain, yielding the strongest overall performance and suggesting that step-level supervision on tool use is more effective than purely outcome-based rewards for fostering robust agentic visual reasoning.

\section{Related Work}

\subsection{Visual Reasoning and Tool Use}
A line of work explores agentic visual reasoning by composing image operations or tools, from VisProg \cite{gupta2023visual} and ViperGPT \cite{suris2023vipergpt}, which express complex visual tasks as executable programs over image operators or a unified Python API, to MM-ReAct \citep{yang2023mm} and Visual Sketchpad~\citep{hu2024visualsketchpad}, which prompt VLMs to interleave perception, search, and calculation while editing intermediate sketches as a visual chain-of-thought. More recent methods, MathCoder-VL \cite{wang2025mathcoder} trains models on large image--code corpora to jointly perceive images and write executable reasoning code. Pixel-Reasoner \cite{su2025pixelreasoner}, DeepEyes \cite{zheng2025deepeyes}, ViRFT \cite{liu2025visualarft}, and OpenThinkIMG \cite{su2025openthinkimg} apply reinforcement learning improve long-horizon multimodal reasoning, while VISTA \cite{huang2025visualtoolagent} and Thyme \cite{zhang2025thyme} investigate more flexible tool interfaces and code-based tool abstractions. Our work builds on these tool-based visual agents by exposing their limitations, which motivates a new focus on evaluating and training for faithful visual tool outputs.


\subsection{Chain-of-Thought Faithfulness}
Early RLHF systems such as InstructGPT \cite{ouyang2022training} relied on PPO \cite{schulman2017proximal} with sparse outcome rewards, suffering from credit-assignment limitations. Subsequent work showed that process reward models with step-level feedback can substantially outperform pure outcome supervision on reasoning tasks \cite{lightman2023let}, while DeepSeekMath \cite{shao2024deepseekmath} demonstrated that GRPO with outcome-only rewards may still induce unfaithful behaviors such as language mixing. More recent frameworks \cite{tao2025hybrid,xu2025hybrid} combine verifiable signals with dense reward-model scores and on-policy distillation of high-reward trajectories to stabilize long-horizon reasoning. Complementary to these textual chain-of-thought methods, we pursue an alternative route that attaches verifiable rewards directly to visual tool outputs, providing step-level supervision that tightly couples the agent’s actions to the underlying evidence.

\section{Conclusion}
In this work, we studied how to train agentic multimodal models that perform faithful visual reasoning. We identified unfaithful behaviors in current open-source models. We then introduced \model, a code-based visual agent trained with Tool-Aware Policy Optimization (\algo), a process-level RL framework that directly rewards faithful visual tool use instead of focusing solely on final-answer accuracy. Empirically, \model attains strong performance and a high rate of faithful tool use across visual search and reasoning benchmarks. These results reveal a key limitation of outcome-only RLVR and highlight the importance of explicitly supervising tool use when training agentic multimodal systems.

\paragraph{Limitations and Future Work} While \model and \algo make progress toward faithful visual tool use, several limitations remain. First, \algo relies on a static judge model, which incurs additional deployment costs; fine-tuning a reward model or self-critic to approximate this judge is an important direction. Second, our current judge is mainly tailored to image-centric operations, such as cropping, and does not provide a general metric for tool-output faithfulness across richer tool types. Extending our framework to broader tool ecosystems (e.g., structured knowledge bases, dynamic web environments) will require designing scalable, verifiable notions of tool faithfulness and corresponding task-specific reward signals.

{
    \small
    \bibliographystyle{ieeenat_fullname}
    \bibliography{main}
}
\clearpage
\setcounter{page}{1}
\maketitlesupplementary
In this supplementary material, we provide (1) details of the faithfulness evaluation protocol and judge prompt (Appendix~\ref{supp:faith}), (2) data preprocessing and
distribution statistics (Appendix~\ref{sec:supp:data}), (3) full benchmark results and additional
tool use analysis (Appendix~\ref{sec:supp:full_result}--\ref{sec:supp:tool_stats}), (4) sandbox implementation details (Appendix~\ref{sec:supp:sandbox}, (5) prompt templates for RL data and reward model (Appendix \ref{sec:supp:prompt}), (6) training dynamics during RL (Appendix \ref{sec:supp:training_dynamics}) and (7) additional qualitative examples (Appendix~\ref{sec:supp:qualitative}).

\section{Faithfulness Evaluation Protocol}
\label{supp:faith}

\begin{prompttemplate}[label={tmpl:crop-judge}]{Faithfulness Judge Prompt}
\textbf{\texttt{[System]}} 

You are evaluating whether an image contains any relevant visual content in the question.

Your task: Determine if the image clearly shows any objects/content mentioned or implied in the question.

Use 1 if:
- Any objects/content mentioned in the question is clearly visible

Use 0 if:
- None of the relevant objects is visible, clearly visible, or identifiable
\par\medskip
\textbf{\texttt{[User]}} 

Question: \{ \}

Does this image clearly show any objects/content mentioned in the question?
\par\medskip
Image. \{ \}
\end{prompttemplate}

\begin{algorithm}[b]
\caption{Faithfulness Evaluation Algorithm}
\label{alg:crop-faithfulness}
\begin{algorithmic}[1]
\Require rollout $\mathcal{R}$, results $\mathcal{A}$
\State Load answer map $\text{Ans}[i] = (\hat y_i, y_i, \text{hit}_i)$ from $\mathcal{A}$
\State Load entries $\{E_1,\dots,E_N\}$ from $\mathcal{R}$
\For{each entry $E_i$}
  \State Extract $q_i$, original image path $I_i$, crop paths $C_i = \{c_{i1},\dots,c_{iK_i}\}$
  \State Infer image index $k$ from $I_i$ and get $\text{hit}_i$ from $\text{Ans}[k]$
  \For{each crop $c_{ij} \in C_i$}
    \State Query GPT-4o with Template~\ref{tmpl:crop-judge} and crop $c_{ij}$
    \State Obtain label $z_{ij} \in \{0,1\}$ (\texttt{contains\_object})
  \EndFor
  \State $\text{any\_crop}_i \gets \mathbb{I}(\exists j : z_{ij} = 1)$
\EndFor
\State $\mathcal{I}_{\text{correct}} \gets \{ i : \text{hit}_i = 1 \}$
\State $\text{faithful} \gets
  \dfrac{\left|\{ i \in \mathcal{I}_{\text{correct}} : \text{any\_crop}_i = 1\}\right|}
        {\left|\mathcal{I}_{\text{correct}}\right|}$
\State \Return faithfulness score $\text{faithful}$
\end{algorithmic}
\end{algorithm}

To quantify action-level faithfulness in visual tool use, we evaluate only the \emph{visual consequences} of each tool call rather than the model’s internal chain-of-thought. Concretely, for every rollout we collect the intermediate cropped images produced during reasoning and, together with the original question, submit each crop to GPT-4o using the \emph{Faithfulness Judge Prompt} (Template~\ref{tmpl:crop-judge}) to decide whether the crop visibly contains the queried object or region. Since these agents are explicitly trained to solve visual search tasks by cropping a small target from a high-resolution image, the presence or absence of the target in the crop provides a direct proxy for whether the tool call follows the intended problem-solving path. The \emph{Faithfulness Evaluation Algorithm} (Algorithm~\ref{alg:crop-faithfulness}) then aggregates these judgments by marking an example as faithful if \emph{any} of its crops is judged to contain the target and computing the proportion of faithful examples among those with correct final answers. This yields a single faithfulness score that measures how often correct predictions are supported by visually grounded tool use, rather than being obtained through shortcuts such as exploiting textual cues.

\section{Data preparation}\label{sec:supp:data}
\point{SFT-data} 
We start from Qwen2.5-VL-7B-Instruct, which has strong coding and perception
capabilities but is not naturally incentivized to write code, inspect visual
evidence, and then answer questions. Direct Zero-RL training from this model
leads to highly unstable updates: the policy either hacks the tool reward when
it is easy to obtain or avoids code entirely once penalties are introduced.
Therefore, we first perform a cold-start SFT stage that explicitly teaches the
model which tools are available and how to use them to support answering.

We build our SFT corpus from Thyme-SFT~\cite{zhang2025thyme}, a curated dataset
distilled from over 4M raw multimodal examples. This dataset contains rich
multi-step traces where the model writes and executes code for image operations
(e.g., cropping high-resolution images, rotation, contrast enhancement) and
mathematical computation. From this corpus, we select single-round and
second-round tool-use examples, resulting in 333K SFT samples.

\begin{figure}[h!]
    \centering
    \begin{subfigure}[t]{\columnwidth}
        \centering
        \includegraphics[width=\linewidth]{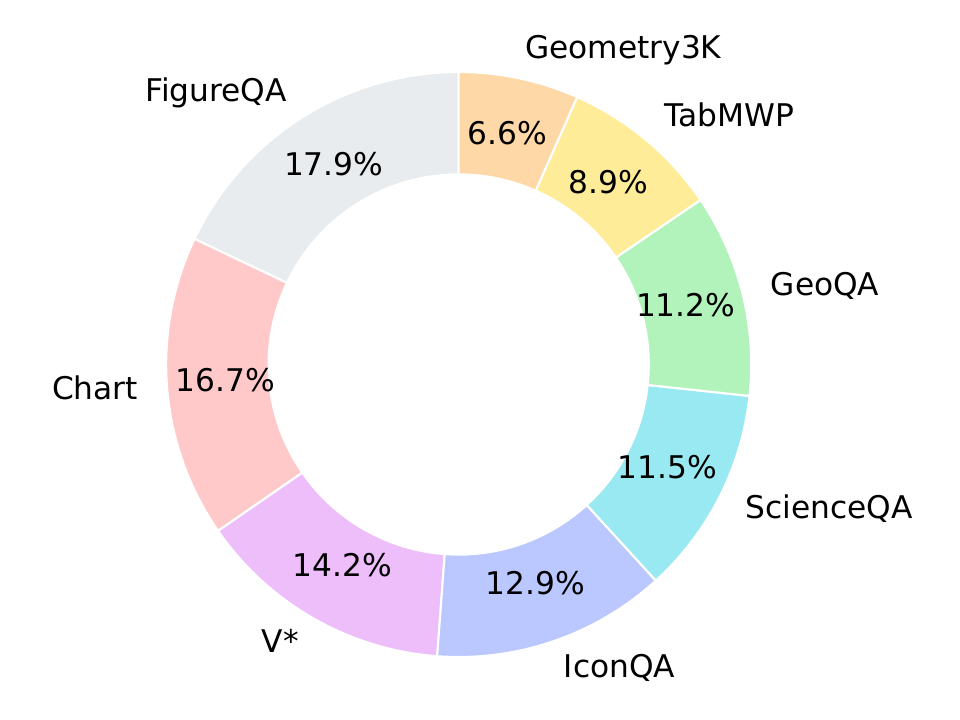}
    \end{subfigure}
    \vspace{-1.5em}
    \caption{RL data distribution.}
    \label{fig:app:rl_data}
\end{figure}

\point{RL-data} 
High-quality RL data is crucial for incentivizing visual reasoning and ensuring
that the model can handle complex tasks. We construct our RL corpus by curating
from open-source datasets such as Thinklite-70K~\cite{wang2025sota} and
DeepEyes-47K~\cite{zheng2025deepeyes}. First, we remove problems that require external
knowledge (e.g., OK-VQA~\cite{marino2019ok}), since these tasks depend on web
search or domain-specific tools rather than visual reasoning. Second, we clean
noisy ground-truth labels via a combination of automatic checks with
Qwen2.5-VL-32B-Instruct and manual verification. Third, we prompt
Qwen2.5-VL-7B-Instruct to answer each question eight times and discard samples
whose empirical accuracy exceeds $0.9$, which reduces near-trivial questions
where all trajectories receive identical rewards and provide little RL signal.

The resulting RL dataset is used exclusively in our TAPO-based RL stage. The
distribution over source datasets is shown in Figure~\ref{fig:app:rl_data},
where FigureQA, ChartQA, V*, and IconQA jointly account for more than half of
the samples.

\section{Full results}
\label{sec:supp:full_result}
\subsection{Full results for main benchmarks}\label{sec:supp:faith}

Table~\ref{tab:perception_results} provides a detailed breakdown on perception-heavy benchmarks, including VLMBlinds, V*, HRBench (4K/8K), and MME-Realworld-Lite. 
We observe that models built upon similar large-scale visual tool-use training data, namely \textit{DeepEyes-7B}, \textit{Thyme-RL-7B}, and our \textit{CodeV} variants, exhibit broadly comparable performance across these tasks. 
This is expected, as these models share overlapping foundations in tool-augmented perception data and visual search supervision.

Within this regime, \model demonstrates a consistent advantage:
\begin{itemize}
    \item On VLMBlinds, \model achieves \underline{46.6}, outperforming both Thyme-RL-7B (38.6) and DeepEyes-7B (41.2), and slightly surpassing GPT-4o (45.1), indicating that incentivizing tool use does not harm low-level perception robustness.
    \item On large-image benchmarks (V*, HRBench-4K/8K), \model matches or exceeds the best-performing tool-based baselines. In particular, it reaches \textbf{84.8} on V* and \textbf{91.0} on HRBench-4K-FSP, while maintaining strong performance on HRBench-8K. 
    \item Compared to its backbone baseline Qwen2.5-VL-7B, CodeV-7B-RL yields consistent improvements across all perception dimensions, with gains ranging from +2.7 on VLMBlinds to +9.8 on V*, as summarized by the $\Delta$ row.
\end{itemize}

These trends suggest that the strong performance of tool-based models on visual search and high-resolution perception tasks is largely grounded in their shared training distributions. Importantly, CodeV further improves upon this foundation through process-level reinforcement learning, leading to more stable and generalizable gains rather than overfitting to specific benchmarks.

\paragraph{Math and Reasoning Benchmarks.}
Table~\ref{tab:math_reason_results} summarizes performance on math-heavy and scientific reasoning benchmarks, including MathVista, CharXiv, MMMU, MathVerse-Mini, and MathVision-Mini.

Compared to proprietary GPT-4o and strong open-source baselines, we observe:
\begin{itemize}
    \item \model achieves the best performance among all models on MathVista (\textbf{71.8}) and substantially narrows the gap with GPT-4o on MathVerse-Mini (49.2 vs. 50.2).
    \item On MathVision-Mini, \model scores \underline{33.6}, closing most of the gap to GPT-4o (35.9) while significantly outperforming all other 7B open-source models by a large margin (e.g., +12.2 over Qwen2.5-VL-7B).
    \item On broader multimodal reasoning tasks like MMMU and CharXiv, CodeV remains competitive, consistently surpassing its 7B peers while approaching the performance of GPT-4o.
\end{itemize}

Taken together, these results show that although our training pipeline is heavily grounded in visual tool use and perception data (similar to Thyme and DeepEyes), the proposed TAPO framework enables a broader transfer of capabilities. Specifically, CodeV not only matches or surpasses peer 7B open-source models across math and reasoning tasks, but also closes much of the remaining gap to GPT-4o on several general-purpose multimodal benchmarks.

\begin{table*}
\centering
\begin{tabular}{lccccccccccc}
\toprule
\multirow{2}{*}{\textbf{Model}} &
\multirow{2}{*}{\textbf{VLMBlind}} &
\multirow{2}{*}{\textbf{V*}} &
\multicolumn{3}{c}{\textbf{HRBench-4K}} &
\multicolumn{3}{c}{\textbf{HRBench-8K}} &
\multicolumn{3}{c}{\textbf{MME-Realworld-Lite}} \\
\cmidrule(lr){4-6} \cmidrule(lr){7-9} \cmidrule(lr){10-12}
 & & &
{\textcolor{gray}{\footnotesize \textbf{ALL}}} &
{\textcolor{gray}{\footnotesize \textbf{FSP}}} &
{\textcolor{gray}{\footnotesize \textbf{FCP}}} &
{\textcolor{gray}{\footnotesize \textbf{ALL}}} &
{\textcolor{gray}{\footnotesize \textbf{FSP}}} &
{\textcolor{gray}{\footnotesize \textbf{FCP}}} &
{\textcolor{gray}{\footnotesize \textbf{All}}} &
{\textcolor{gray}{\footnotesize \textbf{Reasoning}}} &
{\textcolor{gray}{\footnotesize \textbf{Perception}}}
\\
\midrule

\textbf{Model w/o tool use} \\
GPT-4o & 45.1 & 64.4 & 63.1 & 67.8 & 58.5 & 61.3 & 65.3 & \underline{57.3} & 52.0 & 48.3 & 54.4 \\
InternVL3-8B & 44.1 & 76.1 & 70.1 & 79.8 & 61.1 & 69.3 & 78.8 & 59.8 & 48.6 & 44.8 & 51.0  \\
ThinkLite-7B & \textbf{47.4} & 76.4 & 70.4 & 87.0 & 53.8 & 65.9 & 81.8 & 50.0 &39.5& 39.1&39.8\\
Qwen2.5-VL-7B & 43.9 & 75.0 & 68.6 & 82.3 & 55.0 & 63.6 & 75.0 & 52.3 & 44.1 & 37.7 & 48.8 \\
\midrule
\textbf{Model with tool use} \\
DeepEyes-7B & 41.2 & \textbf{84.8} & 73.1 & \textbf{91.0} & 55.3 & 69.1 & \underline{85.3} & 53.0 &  \textbf{53.1} & \textbf{49.2} & \underline{56.2} \\
Pixel-Reasoner-7B & 42.6 & 79.6 & 70.1 & 83.5 & 56.8 & 68.5 & 82.5 & 54.5 & 49.0 & 45.8 & 51.1 \\
Thyme-RL-7B & 38.6 & \underline{82.7} & \textbf{76.8} & \underline{90.5} & \textbf{62.0} & \underline{71.2} & \textbf{86.2} & 57.1 & 51.9 & \underline{48.9} & 55.9 \\

\midrule
\textbf{CodeV-7B-SFT} & 34.4 & 76.4 & 71.5 & 84.8 & 58.3 & 64.5 & 77.8 & 51.3 & -- & -- & -- \\
\textbf{CodeV-7B-RL} & \underline{46.6} & \textbf{84.8} & \underline{76.1} & \textbf{91.0} & \underline{61.3} & \textbf{71.3} & 81.3 & \textbf{60.3} & \underline{52.6} & 47.2 & \textbf{56.3} \\
\textcolor{gray}{$\Delta$ Qwen2.5-VL-7B}
& \textcolor{gray}{+2.7} 
& \textcolor{gray}{+9.8} 
& \textcolor{gray}{+7.5} 
& \textcolor{gray}{+8.7} 
& \textcolor{gray}{+6.3} 
& \textcolor{gray}{+7.7} 
& \textcolor{gray}{+6.3} 
& \textcolor{gray}{+8.0} 
& \textcolor{gray}{+8.5} 
& \textcolor{gray}{+9.5} 
& \textcolor{gray}{+7.5} \\

\toprule
\end{tabular}
\caption{\textbf{Perception benchmarks results.} Comparison of model performance on VLMBlinds, V*, HRBench-4K, HRBench-8K and MME-Realworld-Lite benchmarks (values in \%). 
Best and second-best results in each column are highlighted in \textbf{bold} and \underline{underlined}, respectively.}
\label{tab:perception_results}
\end{table*}

\begin{table*}
\centering
\begin{tabular}{lcccccc}
\toprule
\textbf{Model} & 
\textbf{MathVista} & 
\makecell{\textbf{CharXiv}\\{\textcolor{gray}{\footnotesize \textbf{Reasoning}}}} & 
\makecell{\textbf{CharXiv}\\{\textcolor{gray}{\footnotesize \textbf{Description}}}} & 
\textbf{MMMU} & 
\makecell{\textbf{MathVerse}\\{\textcolor{gray}{\footnotesize \textbf{Mini}}}} & 
\makecell{\textbf{MathVision}\\{\textcolor{gray}{\footnotesize \textbf{Mini}}}} \\
\toprule
\textbf{Model without tool use} \\
GPT-4o & 63.7 & \textbf{45.0} & \textbf{85.4} & \underline{68.7} & \textbf{50.2} & \textbf{35.9} \\

InternVL3-8B & 70.4 & 37.8 & 72.2 & \textbf{70.1} & 40.1 & 26.3 \\
ThinkLite-7B & \underline{71.3} & 38.4 & \underline{75.9} & 57.8 & 48.2 & 25.7 \\
Qwen2.5-VL-7B & 67.9 & 36.3 & 71.8 & 55.2 & 45.5 & 21.4 \\
\midrule
\textbf{Model with tool use} \\
DeepEyes-7B & 68.0 & 36.7 & 70.5 & 52.7 & 45.4 & 26.3 \\
Pixel-Reasoner-7B & 71.2 & 38.8 & 70.7 & 48.7 & 46.9 & 26.3 \\
Thyme-RL-7B & 68.8 & 38.2 & 67.4 & 54.7 & 43.6 & 22.0 \\
\midrule
\textbf{CodeV-7B-SFT }& 68.1 & 32.3 & 70.6 & 48.4 & 44.2 & 23.7 \\
\textbf{CodeV-7B-RL} & \textbf{71.8} & \underline{39.3} & 72.1 & 59.3 & \underline{49.2} & \underline{33.6} \\
\textcolor{gray}{$\Delta$ Qwen2.5-VL-7B}
& \textcolor{gray}{+3.9} 
& \textcolor{gray}{+3.0} 
& \textcolor{gray}{+0.3} 
& \textcolor{gray}{+4.1} 
& \textcolor{gray}{+3.7} 
& \textcolor{gray}{+12.2} \\
\toprule
\end{tabular}
\caption{\textbf{Math and reasoning benchmarks.} Best and second-best results in each column are highlighted in \textbf{bold} and \underline{underlined}, respectively.}
\label{tab:math_reason_results}
\end{table*}

\subsection{Full results on Faithful Tool Use}
\label{sec:supp:faith_discussion}

Table~\ref{tab:faithful_full_results} reports the full results of our faithfulness analysis on V*, HRBench-4K, and HRBench-8K. 
Unlike standard accuracy metrics, this evaluation explicitly measures whether a model’s tool use is \emph{evidence-aligned}, i.e., whether its intermediate visual operations (such as image crops) actually contain the queried region or object, and whether the final answer is correct.

\point{Scaling Difficulty and Resolution}
Another finding is that faithfulness consistently drops as image resolution and task difficulty increase (from V* to HRBench-8K). 
This highlights the inherent challenge of maintaining faithful reasoning in large-scale, high-resolution visual environments, where models must localize fine-grained regions under greater visual complexity. For tasks like this, we believe a multi-agent system that divides the full image into small parts and analyzes them independently is a more practical solution. 
Nevertheless, \model demonstrates significantly better robustness under this increasing difficulty, further validating the effectiveness of our process-level training design.

\begin{table*}
\centering
\begin{tabular}{lcccccc}
\toprule
\multirow{2}{*}{\textbf{Model}} &
\multicolumn{2}{c}{\textbf{V*}} &
\multicolumn{2}{c}{\textbf{HRBench-4K}} &
\multicolumn{2}{c}{\textbf{HRBench-8K}} \\
\cmidrule(lr){2-3} \cmidrule(lr){4-5} \cmidrule(lr){6-7}
 &
{\textcolor{gray}{\footnotesize \textbf{Accuracy}}} &
{\textcolor{gray}{\footnotesize \textbf{Faithful}}} &
{\textcolor{gray}{\footnotesize \textbf{Accuracy}}} &
{\textcolor{gray}{\footnotesize \textbf{Faithful}}} &
{\textcolor{gray}{\footnotesize \textbf{Accuracy}}} &
{\textcolor{gray}{\footnotesize \textbf{Faithful}}} \\
\midrule
DeepEyes-7B        & \textbf{84.8} & 49.7 & 73.1 & 24.7 & 69.1 & 6.7\\
Pixel-Reasoner-7B  & 79.6 & 34.1 & 70.1 & 20.1 & 68.5 & 7.6\\
Thyme-RL-7B           & 82.7 & 7.0 & \textbf{76.8} & 1.0 & 71.2 & 1.2\\
\midrule
\textbf{CodeV-7B-RL}  & \textbf{84.8} & \textbf{68.0 }& 76.1 & \textbf{33.5} & \textbf{71.3} & \textbf{13.3}\\
\bottomrule
\end{tabular}

\caption{\textbf{Faithful results.} \textcolor{gray}{ \textbf{Faithful}} is computed as faithful tool use \textit{AND} correct answer, divided by total number of testing samples. Best result in each column are highlighted in \textbf{bold}.}
\label{tab:faithful_full_results}
\end{table*}

\begin{table*}
\centering
\begin{tabular}{lcccccc}
\toprule
Model &  $\Delta$   & Mask & Noise & Random & Empty \\ \hline
 & Think & 39.7  & 28.7 & 28.3 &16.8\\
Pixel-Reasoner & Action & 18.3 & 17.3 & 6.3 & 2.6 \\
 & Answer & 2.1 & 2.1 & 1.0 & 4.2 \\
\hline
 & Think & 5.8 & 2.6 & 5.2 & 6.3\\
DeepEyes & Action & 0.0 & 0.0 & 0.5 & 0.0\\
 & Answer & 5.2 & 5.8 & 5.2 & 4.7\\
\hline
 & Think & 85.9 & 90.1 & 82.7 & 79.2\\
\textbf{\model-RL} & Action & 48.9 & 50.8 & 44.5 & 24.1\\
 & Answer & 9.6 & 8.0 & 8.9 & 8.4\\
\toprule 
\end{tabular}
\caption{\textbf{Tool-output perturbation analysis.} We evaluate changes in the model’s behavior when all returned tool-output images are perturbed at inference time. \textbf{Mask}: mask out the tool output. \textbf{Noise}: replace the tool output with Gaussian noise. \textbf{Random}: replace the tool output with a non-overlapping random crop. \textbf{Empty}: remove the tool output entirely. We report the percentage of examples with a behavioral change ($\Delta$): \textbf{Think}, changes in reasoning tokens; \textbf{Action}, changes in subsequent tool-use actions (e.g., retrying a crop); and \textbf{Answer}, changes in the final answer.}

\label{reb:perturb}
\end{table*}

\subsection{Full results for ablation study}\label{sec:supp:ablation}
The abalation results reported in Section~\ref{exp:ablation} present a summarized view averaged across benchmarks. 
Tables~\ref{tab:ablation_full_results} and~\ref{tab:ablation_perception_results} provide the complete, per-benchmark breakdown.

From these full results, we observe that individual training stages and reward components affect different benchmarks unevenly: some variants yield gains on specific perception subsets (e.g., HRBench-FSP) or structured reasoning tasks (e.g., CharXiv-Reasoning), but often at the cost of regressions elsewhere. 
In contrast, \model shows consistently strong performance across nearly all perception and reasoning benchmarks, indicating that its advantages are not driven by isolated datasets but reflect improved general robustness.

\begin{table*}
\centering
\begin{tabular}{lccccccccc}
\toprule
\multirow{2}{*}{\textbf{Model}} &
\multirow{2}{*}{\textbf{VLMBlinds}} &
\multirow{2}{*}{\textbf{V*}} &
\multicolumn{3}{c}{\textbf{HRBench-4K}} &
\multicolumn{3}{c}{\textbf{HRBench-8K}} \\
\cmidrule(lr){4-6} \cmidrule(lr){7-9}
 & & &
{\textcolor{gray}{\footnotesize \textbf{ALL}}} &
{\textcolor{gray}{\footnotesize \textbf{FSP}}} &
{\textcolor{gray}{\footnotesize \textbf{FCP}}} &
{\textcolor{gray}{\footnotesize \textbf{ALL}}} &
{\textcolor{gray}{\footnotesize \textbf{FSP}}} &
{\textcolor{gray}{\footnotesize \textbf{FCP}}} \\
\midrule
\textbf{Training stage} \\
Qwen2.5-VL-7B      & 43.9 & 75.0 & 68.6 & 82.2 & 55.0 & 63.6 & 75.0 & 52.2 \\
Zero-RL            & 46.6 & 78.5 & 73.0 & 89.0 & 57.0 & 69.9 & 84.2 & 55.5 \\
Cold-start SFT     & 34.4 & 76.4 & 71.5 & 84.8 & 58.2 & 64.5 & 77.8 & 51.2 \\
\toprule
\textbf{Reward design}\\
\ \ \ Accuracy only   & 43.8 & 82.7 & 74.9 & 89.8 & 60.0 & 68.8 & 82.8 & 54.8 \\
+ Consistency         & 45.0 & 83.2 & 74.4 & 87.5 & 61.3 & 67.9 & 79.0 & 56.8 \\
+ GPT-5-nano Judge    & 44.7 & 84.3 & 74.8 & 89.8 & 59.8 & 71.1 & 85.2 & 57.0 \\
\textbf{\model-7B-RL} & \textbf{46.7} & \textbf{84.8} & \textbf{76.1} & \textbf{91.0} & \textbf{61.3} & \textbf{71.2} & \textbf{81.2} & \textbf{60.2} \\
\bottomrule
\end{tabular}
\caption{Ablation study of model performance on Math and reasoning benchmarks (values in \%).}
\label{tab:ablation_full_results}
\end{table*}

\begin{table*}
\centering
\begin{tabular}{lcccccc}
\toprule
\textbf{Model} & 
\textbf{MathVista} & 
\makecell{\textbf{CharXiv}\\{\textcolor{gray}{\footnotesize \textbf{Reasoning}}}} & 
\makecell{\textbf{CharXiv}\\{\textcolor{gray}{\footnotesize \textbf{Description}}}} & 
\textbf{MMMU} & 
\makecell{\textbf{MathVerse}\\{\textcolor{gray}{\footnotesize \textbf{Mini}}}} & 
\makecell{\textbf{MathVision}\\{\textcolor{gray}{\footnotesize \textbf{Mini}}}} \\
\toprule
\textbf{Training stage} \\
Qwen2.5-VL-7B             & 67.9 & 36.3 & 71.8 & 55.2 & 45.5 & 21.4 \\
Zero-RL             & 69.8 & 38.2 & 71.5 & \textbf{59.3} & 48.2 & 29.9 \\
Cold-start SFT                 & 68.1 & 32.3 & 70.6 & 48.4 & 44.2 & 23.7 \\
\toprule
\textbf{Reward design} \\
\ \ \ Accuracy only       & 69.0 & 39.0 & 71.4 & 52.8 & 47.9 & 27.3 \\
+ Consistency       & 69.3 & 36.6 & 70.5 & 52.0 & 48.0 & 26.0 \\
+ GPT-5-nano Judge  & \textbf{72.4} & \textbf{39.6} & \textbf{72.5} & 56.7 & \textbf{49.7} & 23.7 \\
\textbf{\model-7B-RL}          & 71.8 & 39.3 & 72.0 & \textbf{59.3} & 49.2 & \textbf{33.6 }\\
\bottomrule
\end{tabular}
\caption{Ablation study of model performance on VLMBlinds, V*, and HR benchmarks (values in \%).}
\label{tab:ablation_perception_results}
\end{table*}

\begin{figure*}[h!]
    \centering
    \includegraphics[width=\textwidth]{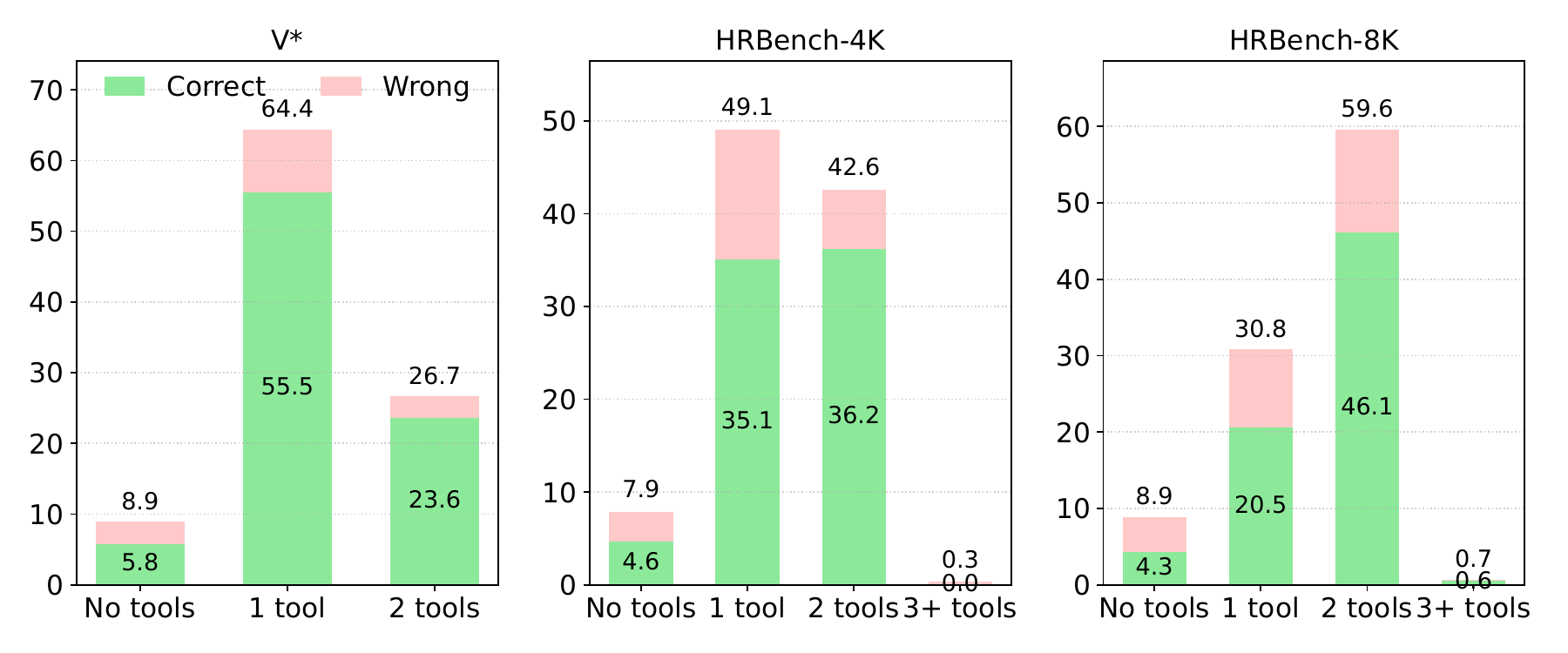}
    \caption{\textbf{Tool-use frequency and accuracy of \model.} 
    For each dataset (V*, HRBench-4K, HRBench-8K), we bucket examples by the number of tool calls in a trajectory and plot the fraction of all test examples in that bucket, split into correct (green) and incorrect (red) answers. 
    Numbers above each bar show the total fraction of examples in that bucket, and numbers inside the green bars show the fraction of \emph{all} examples that are both correct and use that number of tools.}
    \label{fig:supp:tool_use}
\end{figure*}

\section{Perturbation Analysis}
\label{sec:supp:perturb}
Producing a correct crop does not necessarily imply that the model actually uses the tool output. To further assess faithfulness, we perform a \emph{tool-output perturbation} analysis. Specifically, on the V* benchmark, we perturb all returned tool output at inference time and  measure whether the model’s behavior changes. We consider four perturbation types: 
\begin{itemize}
    \item \textbf{Mask}: mask out the tool-output images.
    \item \textbf{Noise}: replace the tool-output images with Gaussian noise.
    \item \textbf{Random}: replace the tool-output images with a non-overlapping random crop.
    \item \textbf{Empty}: remove the tool-output image entirely.
\end{itemize}

We report the percentage of trajectories that exhibit a behavioral change along three dimensions:
\begin{itemize}
    \item \textbf{Think}: changes in reasoning tokens.
    \item \textbf{Action}: changes in subsequent tool-use actions (e.g., retry cropping).
    \item \textbf{Answer}: changes in final answer.
\end{itemize}
Higher values indicate that the model is more sensitive to perturbations of the tool output, suggesting that its reasoning, tool-use behavior, or final prediction depends more strongly on the returned visual evidence.

Table \ref{reb:perturb} shows that Pixel-Reasoner and DeepEyes are largely insensitive to perturbed tool outputs. In contrast, \model exhibits substantially larger behavior changes in \textbf{Think} and \textbf{Action}, indicating that its reasoning process and tool-use decisions are more strongly conditioned on the returned tool outputs. For the final answer, \model is also noticeably more sensitive, with an answer-change rate of around 8--10\%, compared with roughly 1--5\% for the baselines. 

A closer examination of \model trajectories suggests that, under perturbation, the model often switches to alternative reasoning paths (e.g., text-only reasoning or direct producing an answer instead). As a result, the final naswer frequently remains unchanged even when the internal reasoing and action sequence differ. Overall, these results indicate that \model genuinely relies on tool outputs during inference, and this behavior emerges from TAPO's simple process-aware objective, despite not being explicitly trained with perturbation-based supervision.

\section{Tool Use analysis}
\label{sec:supp:tool_stats}

Figure~\ref{fig:supp:tool_use} analyzes how often \model invokes tools and how this correlates with accuracy on V*, HRBench-4K, and HRBench-8K. 
For each dataset, we bucket episodes by the number of tool calls in a trajectory (“No tools”, “1 tool”, “2 tools”, “3+ tools”) and report (i) the fraction of all test examples in each bucket and (ii) the accuracy within that bucket.
Concretely, the green portion of each bar corresponds to
\(\text{(\% of all examples using this number of tools)} \times \text{accuracy in this bucket}\),
so, for example, the “2 tools” bar on HRBench-8K indicates that \model answers correctly on 46.1\% of all test examples using exactly two tool calls.

Across all three benchmarks, most responses use either one or two tool calls, with 3+ tools being extremely rare ($\leq$1\% of examples). 
On V*, performance is dominated by single-tool trajectories, whereas on HRBench-4K and HRBench-8K the mass shifts toward two-tool trajectories, reflecting the need for slightly deeper visual interaction at higher resolutions. 
In all cases, zero-tool answers are less accurate than one–two tool trajectories, and additional tool calls beyond two do not yield clear gains, suggesting that \model learns to use a small number of focused tool calls rather than over-invoking tools.

\section{Python Sandbox Design}\label{sec:supp:sandbox}
We execute all model-generated Python code inside a dedicated sandbox with three goals: (i) security and isolation, (ii) reduced coding burden, and (iii) robust error handling.

\paragraph{Security and isolation}
The sandbox isolates execution from the host system and routes all file I/O into a controlled workspace, so that code cannot affect external files or processes. We statically scan generated code for dangerous filesystem operations (e.g., deleting, moving, or renaming files) and block execution if they are detected. Each tool call is further constrained by a strict wall-clock time limit; exceeding this limit raises a timeout error and aborts the run, preventing malformed programs from stalling training.

\paragraph{Usability and code normalization}
To make code execution more forgiving, the sandbox automatically handles many routine details. Before execution, we normalize working directories and relative paths, auto-format and patch minor issues (e.g., indentation or simple I/O omissions), predefine common variables such as the input image path and loaded image, and pre-import frequently used libraries for image I/O and basic vision operations. We use lightweight static analysis to infer cropping boxes and clamp coordinates to valid image boundaries, avoiding crashes from slightly out-of-range indices. Across multiple tool calls in the same trajectory, we maintain imports and variable definitions so that later code segments can reuse earlier results.

\paragraph{Visualization and artifacts as observations}
Instead of requiring the model to manage filenames or GUI windows, the sandbox tracks all newly created artifacts in the workspace as potential observations. We monkey-patch plotting backends (e.g., intercepting calls to \texttt{plt.show()}) so that figures are captured as images and returned as part of the observation $o_t$, turning visualizations into explicit evidence for the agent rather than ephemeral side effects.

\paragraph{Error handling and iterative repair}
Each tool call is wrapped with exception handling. If execution fails, the resulting error message is returned to the model, which can then decide whether to revise and regenerate the code or bypass the tool and continue reasoning. This design exposes a large but usable action space: the policy can freely “think with code and images” while the sandbox guarantees safe, consistent, and debuggable execution.

\clearpage
\section{RL Prompt Design}\label{sec:supp:prompt}

\begin{prompttemplate}[label={tmpl:rl_data_prompt}]{RL Data Prompt}
\textbf{\texttt{[System]}} 

You are a helpful assistant.
\par\smallskip
**Goal**:\\
    - Answer the user's question based on the provided image and question.\\ 
    - You can optionally generate and execute Python code to help analyze or process the image before answering.
\par\smallskip
**Python execution rules**:\\
    - The Python code will be executed by an external sandbox, and its output will be returned to you in the format <sandbox\_output>...</sandbox\_output>. \\
    - Use sandbox output to decide whether to answer the question or run another round of Python code. \\
    - For image operations, load the image and then process it (crop, resize, rotate, adjust contrast).\\
    - Save any processed images and print the saved filename. \\
    - For calculation, define all variables and print output. \\
    - Python code must be wrapped in the following block: \\
<code> \\
\verb|```| python \\
\# your python code here\\
\verb|```|\\
</code>
\par\medskip
\textbf{\texttt{[User]}} 

Image: \{ \}
Question: \{ \}

\#\#\# User Image Path:** \verb|"|\{file\_path\} \verb|"|\\
\#\#\# User Image Size:** \verb|"|\{image\_x\}$\times$\{image\_y\}\verb|"|
\#\#\# **Output Format (strict adherence required):**

<think>Your detailed reasoning process, including any <code> </code>, should go here.</think>\\
<answer>Answer to the user's question.</answer>
\end{prompttemplate}

\begin{prompttemplate}[label={tmpl:tapo_vs_prompt}]{Reward model prompt for perception}
\textbf{\texttt{[System]}} 
You are an expert judge to score a Agent Response's tool use quality.

Agent generate code to process the image, processed image will appear after \textbf{Agent Generated Images}

- Score = 1 if \textbf{Agent Generated Images} clearly contains at least one object mentioned in the question.

- Score = 0.5 if \textbf{Agent Generated Images} shows a PARTIAL match.

- Score = 0.25 if \textbf{Agent Generated Images} doesn't matches.

\textbf{\texttt{[User]}} 

\#\# Question: \{ \}

**Agent Generated Images**: \{ \}
\end{prompttemplate}
\section{Training Dynamics}\label{sec:supp:training_dynamics}
\begin{figure}[h]
    \centering
    \begin{subfigure}[t]{\columnwidth}
        \centering
        \includegraphics[width=\linewidth]{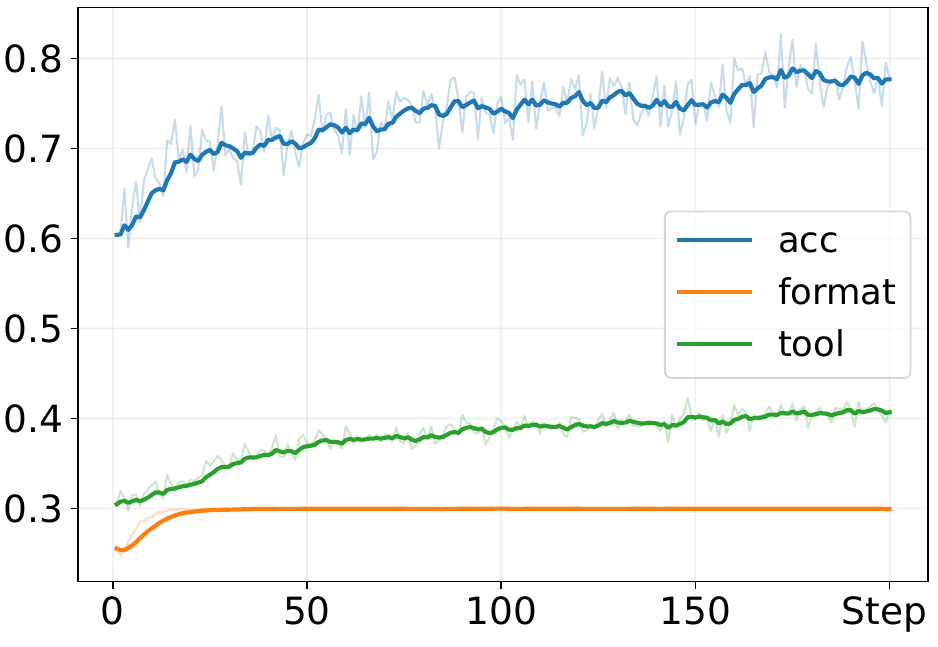}
    \end{subfigure}
    \vspace{-1.5em}
    \caption{Reward score for accuracy, format and tool use.}\label{fig:reward_score}
    \vspace{-1.5em}
\end{figure}

\begin{figure}[h]
    \centering
    \begin{subfigure}[t]{\columnwidth}
        \centering
        \includegraphics[width=\linewidth]{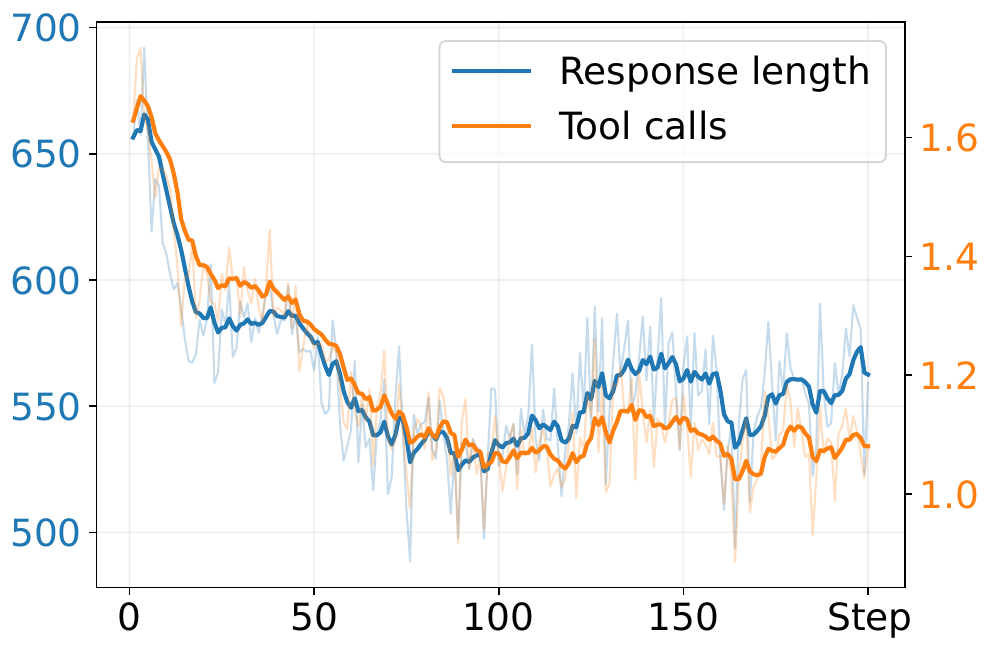}
    \end{subfigure}
    \vspace{-1.5em}
    \caption{Response length and tool calls during \algo .}\label{fig:resp_len_tool_call}
    \vspace{-1.5em}
\end{figure}

During TAPO reinforcement learning, we observe stable and monotonic improvement in all three critic rewards (Figure~\ref{fig:reward_score}). The format reward quickly saturates near its max value (0.3), while the accuracy and tool-consistency rewards continue to rise throughout training, indicating that the policy steadily improves both answer correctness and how tools are used. In parallel, the average response length and the mean number of tool calls per step gradually decrease (Figure~\ref{fig:resp_len_tool_call}), even as rewards increase, suggesting that TAPO discourages reward hacking via overly long chains or gratuitous tool invocations and instead guides the model toward shorter, more targeted, and faithful visual reasoning with tool use.



\clearpage
\newpage

\section{Qualitative Examples}
\label{sec:supp:qualitative}
We provide two faithful and correct examples and two unfaithful examples. These results shows the exact rollouts interleaving text, image, code and tool response.
\subsection{Faithful Example 1}
\begin{promptbox}[User Prompt]
\textbf{Question:} What is the color of the slippers?  

\textbf{Options:}  
A. black \quad B. red \quad C. blue \quad D. pink  

Please select the correct answer from the options above.  

\textbf{User Image Path:} \texttt{"12.jpg"}  

\textbf{User Image Size:} \texttt{"2251x1500"}  

\textbf{Output Format (strict adherence required):}

\verb|<think>| Your detailed reasoning process, including any \verb|<code>|, should go here. \verb|</think>|  

\verb|<answer>| Your final answer to the user's question goes here. \verb|</answer>|
\end{promptbox}
\noindent
    \begin{myframe}
        \includegraphics[width=\linewidth]{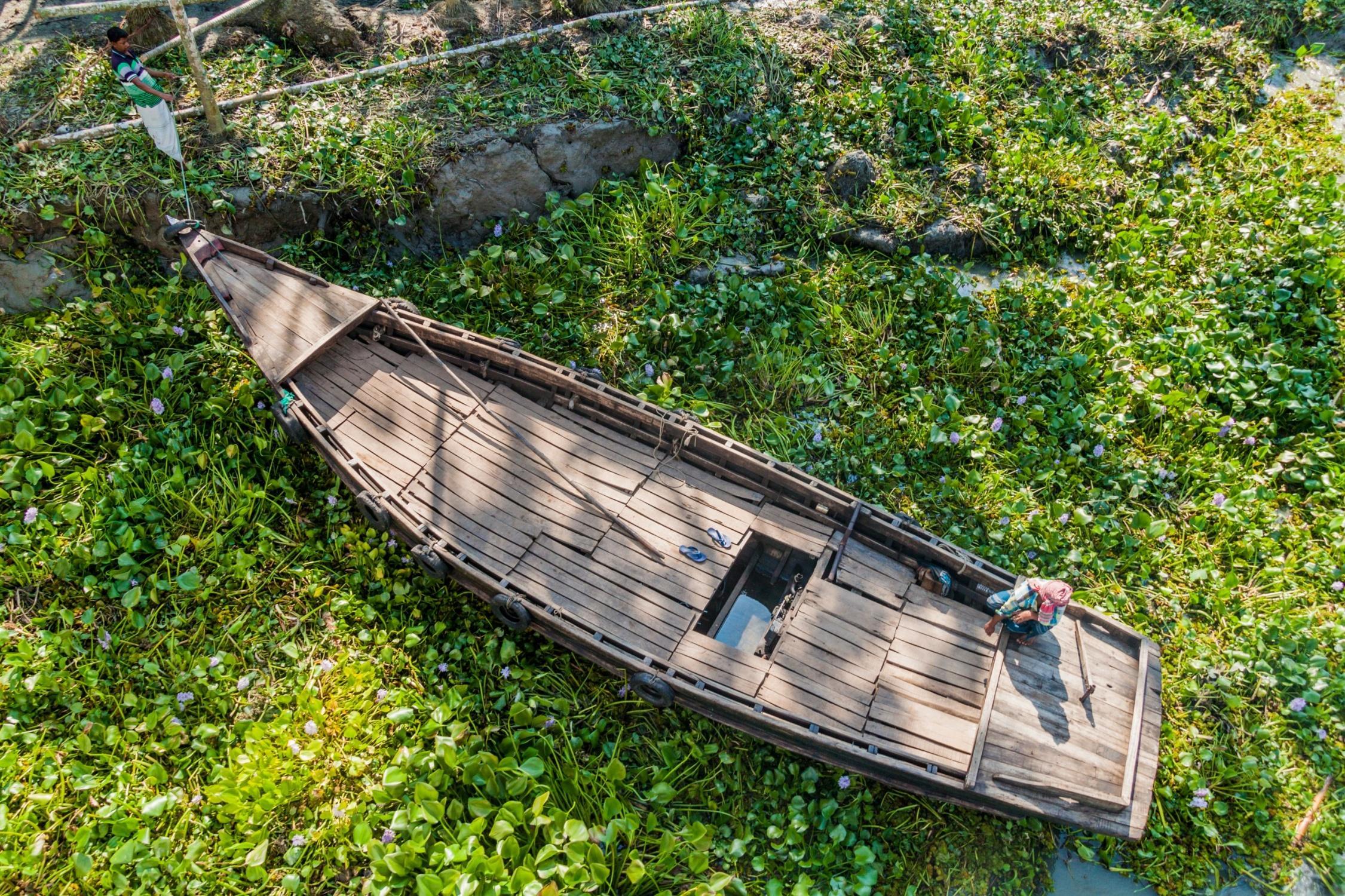}
    \end{myframe}
\label{app:success_1}

\begin{thinkcodebox}[Assistant]
{\bfseries\ttfamily \string<think\string>} The task involves identifying the color of the slippers on the boat. The slippers are located near the center of the boat, and the image is quite wide, making the slippers relatively small. To accurately determine their color, it is necessary to zoom in on the specific region where the slippers are located. This will enhance the visibility of the slippers and allow for a more precise identification of their color.

Here is the Python code to crop and zoom in on the slippers:


{\bfseries\ttfamily \string<code\string>}\par
\lstset{style=modelcode}
\begin{lstlisting}
import cv2
import numpy as np
import os
import random

# Load the image
image_path = "12.jpg"
image = cv2.imread(image_path)

# Define the coordinates for cropping the region around the slippers
x1, y1, x2, y2 = 1100, 850, 1300, 950

# Crop the image
cropped_image = image[y1:y2, x1:x2]

# Apply a zoom factor to enhance the visibility of the slippers
zoom_factor = 2
resized_image = cv2.resize(cropped_image, None, fx=zoom_factor, fy=zoom_factor, interpolation=cv2.INTER_LINEAR)

# Save the processed image with a simple filename
processed_path = "processed_1.jpg"
cv2.imwrite(processed_path, resized_image)

# Print the path of the saved cropped image
print(processed_path)
\end{lstlisting}
{\bfseries\ttfamily \string</code\string>}
{\bfseries\ttfamily \string</think\string>}\par
\end{thinkcodebox}

\bigskip

\begin{sandboxbox}[Sandbox Output]
\centering
\includegraphics[width=0.6\linewidth]{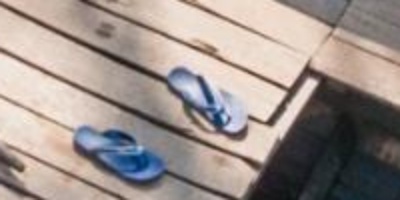}
\end{sandboxbox}

\bigskip

\begin{answerbox}[Assistant]

{\bfseries\ttfamily \string<think\string>}Upon examining the processed image, the slippers are clearly visible near the center of the boat. The enhanced visibility confirms that the slippers are blue in color. This detailed view allows for a precise identification of their color, addressing the original user question effectively.
{\bfseries\ttfamily \string</think\string>}\par

{\bfseries\ttfamily \string<answer\string>} C. blue {\bfseries\ttfamily \string</answer\string>}
\end{answerbox}

\newpage
\subsection{Faithful Example 2}
\begin{promptbox}[User Prompt]
\textbf{Question:} What is the color of the candles?  

\textbf{Options:}  
A. red\quad B. yellow\quad C. gray\quad D. white\

Please select the correct answer from the options above.  

\textbf{User Image Path:} \texttt{"6.jpg"}  

\textbf{User Image Size:} \texttt{"1998x1500"}  

\textbf{Output Format (strict adherence required):}

\verb|<think>| Your detailed reasoning process, including any \verb|<code>|, should go here. \verb|</think>|  

\verb|<answer>| Your final answer to the user's question goes here. \verb|</answer>|
\end{promptbox}

\begin{myframe}
    \includegraphics[width=\linewidth]{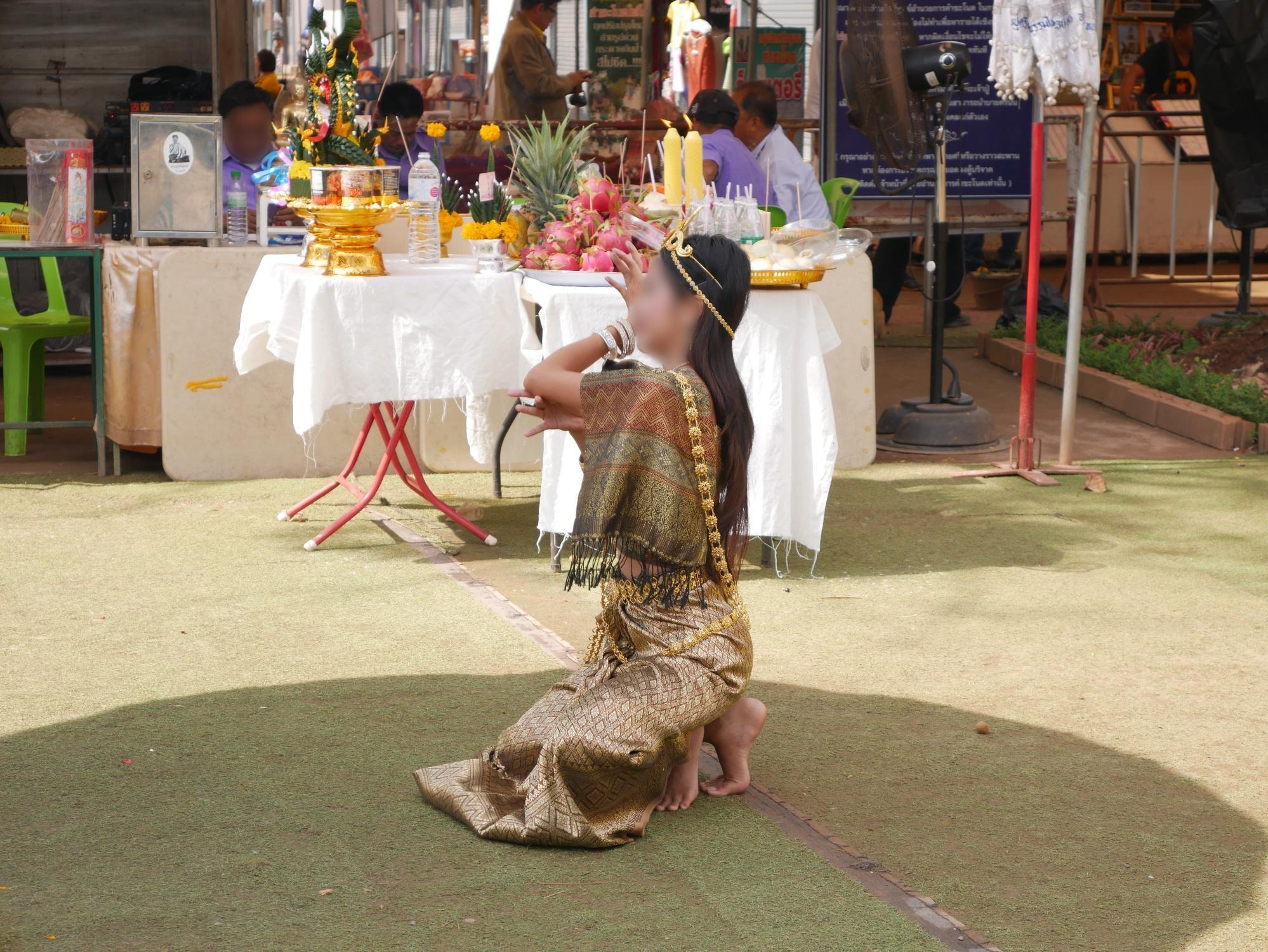}
\end{myframe}

\label{app:success_2}

\begin{thinkcodebox}[Assistant]
{\bfseries\ttfamily \string<think\string>}
The task is to determine the color of the candles in the image. The candles are located on the table in the background, and they are relatively small compared to the overall scene. To accurately identify their color, it is necessary to zoom in on the area around the candles. This will help in clearly distinguishing the color of the candles from the other objects in the image. The coordinates for cropping should focus on the table area where the candles are placed.\\ Here is the Python code to crop and zoom in on the relevant area:

{\bfseries\ttfamily \string<code\string>}\par
\lstset{style=modelcode}
\begin{lstlisting}
import cv2
import os
import numpy as np
from uuid import uuid4

# Load the image
image_path = "6.jpg"
image = cv2.imread(image_path)

# Define the coordinates for cropping (x1, y1, x2, y2)
x1, y1, x2, y2 = 950, 150, 1200, 400

# Crop the image\ncropped_image = image[y1:y2, x1:x2]

# Apply zoom (resize the cropped image to a larger size)
zoom_factor = 2
zoomed_image = cv2.resize(cropped_image, None, fx=zoom_factor, fy=zoom_factor, interpolation=cv2.INTER_LINEAR)

# Save the processed image with a simple filename\nprocessed_path = \"processed_1.jpg\"
cv2.imwrite(processed_path, zoomed_image)

# Print the path of the saved cropped image
print(processed_path)
\end{lstlisting}
{\bfseries\ttfamily \string</code\string>}
{\bfseries\ttfamily \string</think\string>}\par
\end{thinkcodebox}

\bigskip

\begin{sandboxbox}[Sandbox Output]
\centering
\includegraphics[width=0.4\linewidth]{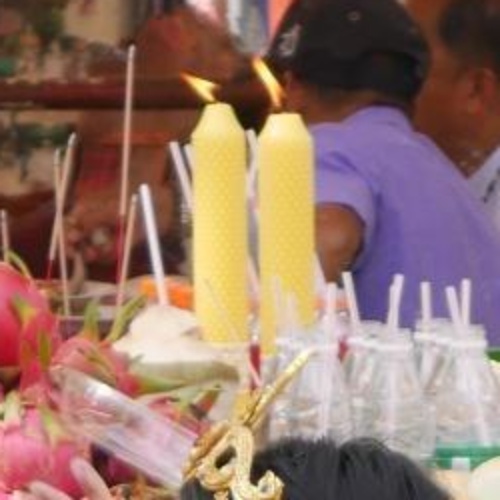}
\end{sandboxbox}

\bigskip

\begin{answerbox}[Assistant]
{\bfseries\ttfamily \string<think\string>}Upon examining the cropped and zoomed image, the candles on the table in the background are clearly visible. The candles appear to be yellow in color. This observation aligns with the options provided in the question. The zoomed-in view effectively highlights the candles, making it easier to distinguish their color from the surrounding objects. Therefore, the correct answer is that the candles are yellow.
{\bfseries\ttfamily \string</think\string>}

{\bfseries\ttfamily \string<answer\string>} B. yellow {\bfseries\ttfamily \string</answer\string>}
\end{answerbox}

\newpage
\subsection{Unfaithful Example 1}
\begin{promptbox}[User Prompt]
\textbf{Question:} Is the mailbox on the left or right side of the woman?  

\textbf{Options:}  
A. In front
B. Left
C. Right
D. Directly above

Please select the correct answer from the options above. 

\textbf{User Image Path:} \texttt{"10.jpg"}  

\textbf{User Image Size:} \texttt{"4992x7680"}  

\textbf{Output Format (strict adherence required):}

\verb|<think>| Your detailed reasoning process, including any \verb|<code>|, should go here. \verb|</think>|  

\verb|<answer>| Your final answer to the user's question goes here. \verb|</answer>|
\end{promptbox}

\begin{myframe}
    \includegraphics[width=\linewidth]{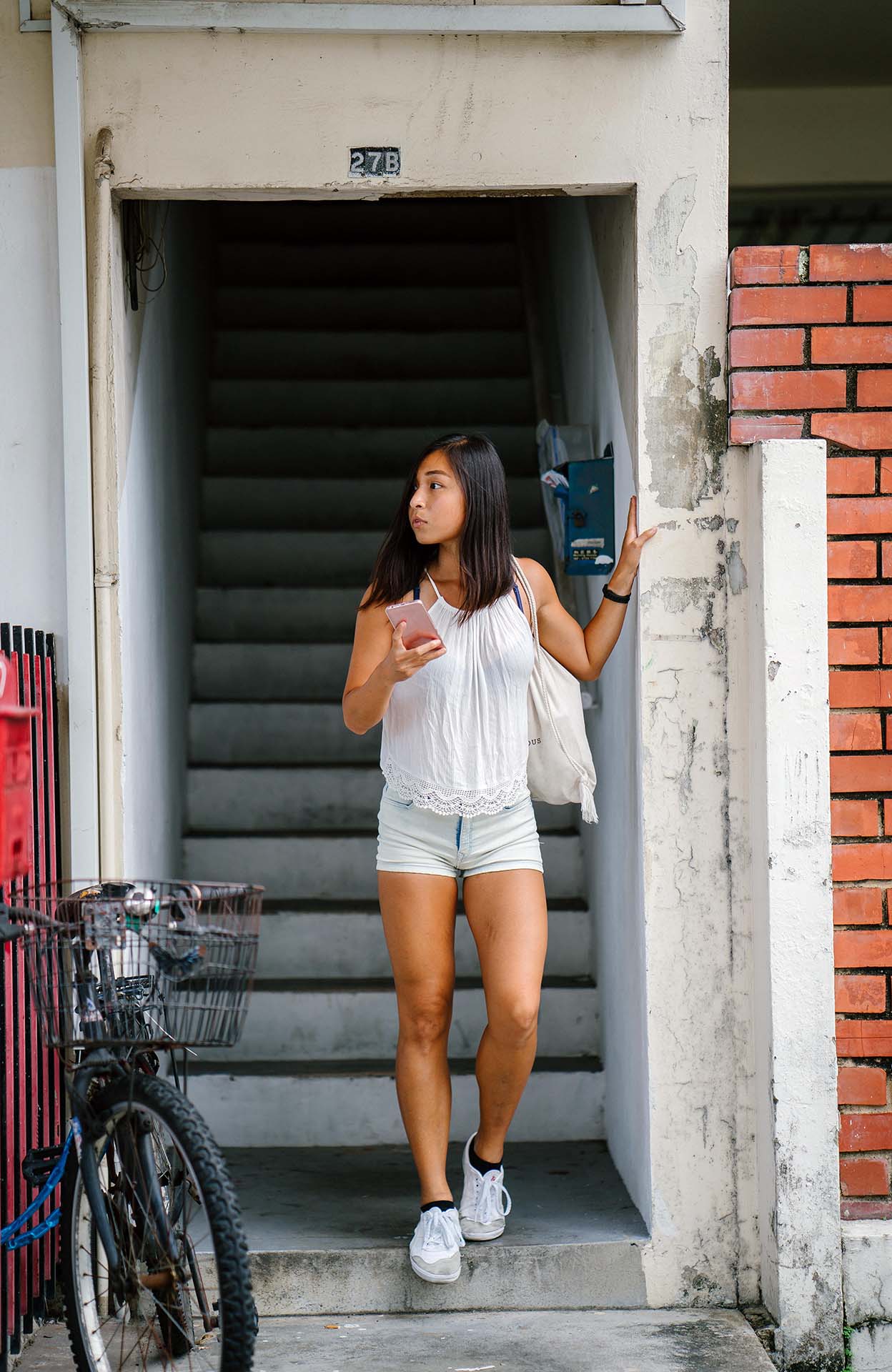}
\end{myframe}

\label{app:success_3}

\begin{thinkcodebox}[Assistant]
{\bfseries\ttfamily \string<think\string>}
To determine the position of the mailbox relative to the woman, we need to focus on the area around her and the mailbox. The mailbox is located near the wall to her right. Cropping and zooming in on this specific region will help clarify its position relative to the woman. This will ensure that we can accurately identify whether it is on her left, right, or in front of her.

The task involves analyzing the relative position of the mailbox concerning the woman. To improve visibility, I will crop the region around the woman and the mailbox. The coordinates appear to be near the right side of the woman, close to the wall. I will refine this area for better focus and clarity.

Here is the Python code to crop and zoom in on the relevant area:

{\bfseries\ttfamily \string<code\string>}\par
\lstset{style=modelcode}
\begin{lstlisting}
import os
import random
from PIL import Image

# Load the image
image_path = "10.jpg"
image = Image.open(image_path)

# Define the cropping coordinates (x1, y1, x2, y2)
x1, y1, x2, y2 = 1600, 1000, 2500, 3000

# Crop the image
cropped_image = image.crop((x1, y1, x2, y2))

# Save the cropped image with a simple filename
processed_path = "cropped_1.jpg"
cropped_image.save(processed_path)

# Print the path of the saved cropped image
print(processed_path)
\end{lstlisting}
{\bfseries\ttfamily \string</code\string>}
{\bfseries\ttfamily \string</think\string>}\par
\end{thinkcodebox}

\bigskip

\begin{sandboxbox}[Sandbox Output]
\centering
\includegraphics[width=0.4\linewidth]{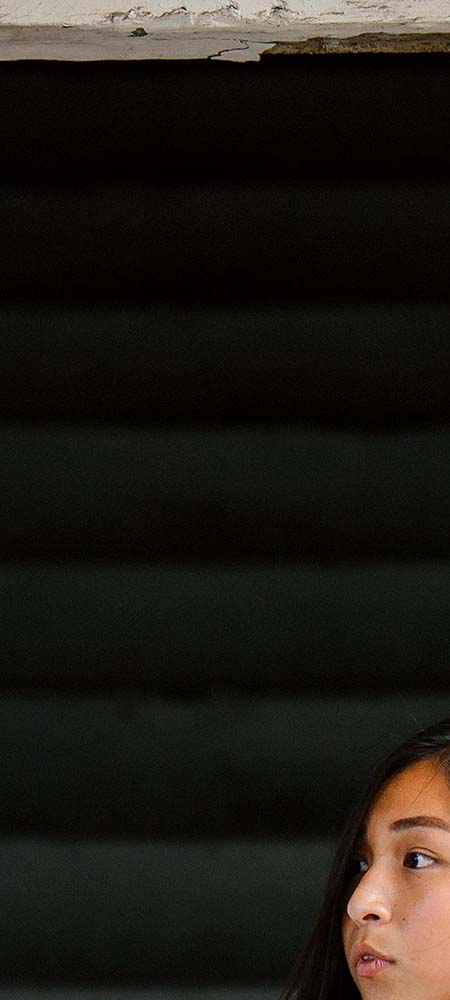}
\end{sandboxbox}

\bigskip

\begin{thinkcodebox}[Assistant]
{\bfseries\ttfamily \string<think\string>}
The initial cropping did not effectively capture the mailbox and its position relative to the woman. The cropped image only shows a close-up of the woman's face and the wall, missing the mailbox entirely. To address this, the cropping range needs to be adjusted to include both the woman and the mailbox.

Here is the corrected cropping code to ensure the mailbox and its position relative to the woman are visible:

{\bfseries\ttfamily \string<code\string>}\par
\lstset{style=modelcode}
\begin{lstlisting}
import os
import random
from PIL import Image

# Load the image
image_path = "10.jpg"
image = Image.open(image_path)

# Define the cropping coordinates (x1, y1, x2, y2)
x1, y1, x2, y2 = 1000, 500, 2800, 5000

# Crop the image
cropped_image = image.crop((x1, y1, x2, y2))

# Save the cropped image with a simple filename
processed_path = "cropped_1.jpg"
cropped_image.save(processed_path)

# Print the path of the saved cropped image
print(processed_path)
\end{lstlisting}
{\bfseries\ttfamily \string</code\string>}
{\bfseries\ttfamily \string</think\string>}\par
\end{thinkcodebox}

\bigskip
\begin{sandboxbox}[Sandbox Output]
\centering
\includegraphics[width=0.4\linewidth]{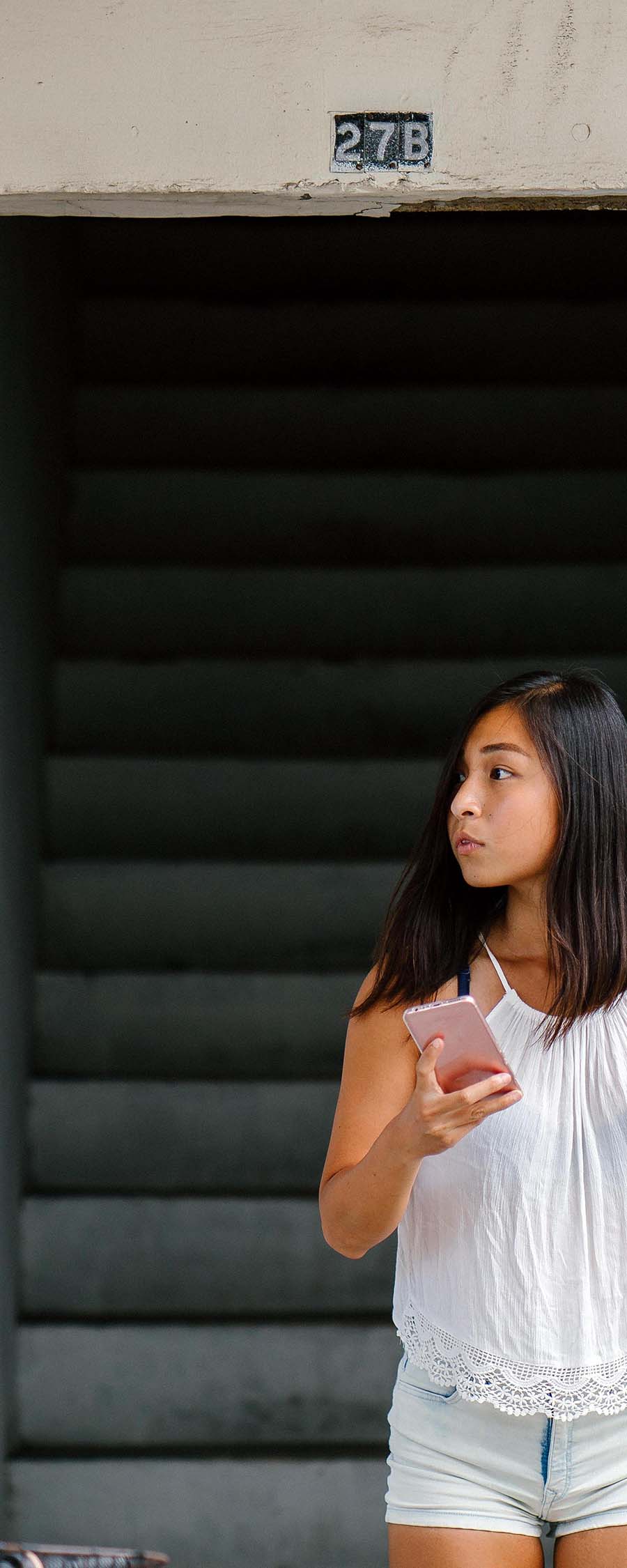}
\end{sandboxbox}
\begin{answerbox}[Assistant]
{\bfseries\ttfamily \string<think\string>}The second attempt at cropping successfully includes both the woman and the mailbox. The mailbox is clearly visible on the right side of the woman, confirming its position. Therefore, the mailbox is on the right side of the woman.
{\bfseries\ttfamily \string</think\string>}

{\bfseries\ttfamily \string<answer\string>} C. Right{\bfseries\ttfamily \string</answer\string>}
\end{answerbox}

\newpage
\subsection{Unfaithful Example 2}
\begin{promptbox}[User Prompt]
\textbf{Question:} Where is the person standing in relation to the recycle bin?

\textbf{Options:}  \\
A. In front of the recycle bin\\
B. To the right of the recycle bin\\
C. To the left of the recycle bin\\
D. Behind the recycle bin\\
Please select the correct answer from the options above. 

\textbf{User Image Path:} \texttt{"48.jpg"}  

\textbf{User Image Size:} \texttt{"7680x4992"}  

\textbf{Output Format (strict adherence required):}

\verb|<think>| Your detailed reasoning process, including any \verb|<code>|, should go here. \verb|</think>|  

\verb|<answer>| Your final answer to the user's question goes here. \verb|</answer>|
\end{promptbox}

\begin{myframe}
    \includegraphics[width=\linewidth]{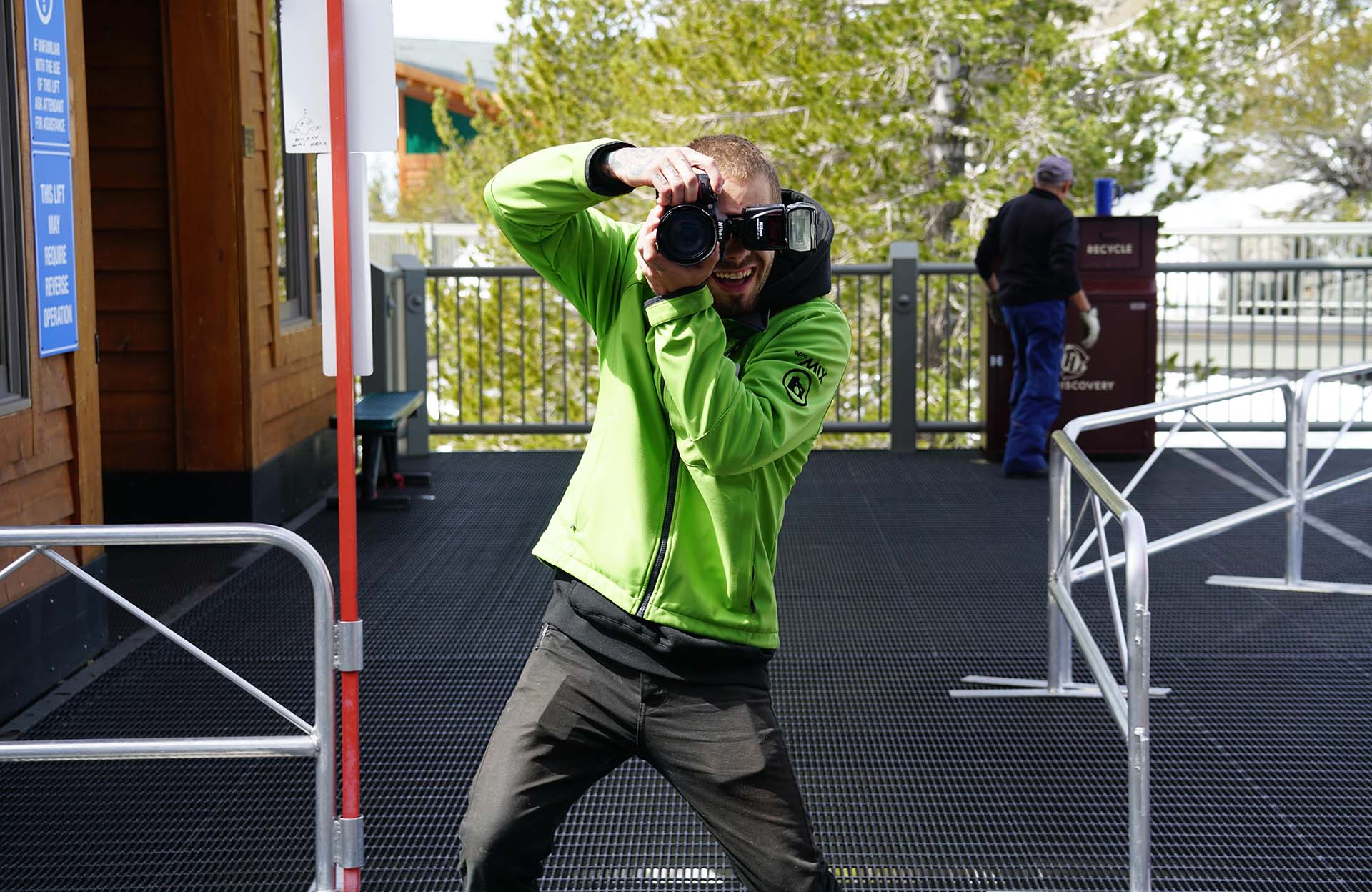}
    
\end{myframe}

\label{app:success_4}

\begin{thinkcodebox}[Assistant]
{\bfseries\ttfamily \string<think\string>}
The task involves determining the relative position of the person in relation to the recycle bin. The person is clearly visible in the foreground, and the recycle bin is in the background to the right. To better analyze the spatial relationship, cropping the image to focus on the area around the person and the recycle bin will enhance clarity. This will help in accurately assessing whether the person is in front of, to the right of, to the left of, or behind the recycle bin.

The coordinates for cropping should be around the person and the recycle bin. Based on the image, the person is positioned roughly in the center-right, and the recycle bin is further to the right in the background. Cropping the image to focus on this area will provide a clearer view.

Here is the Python code to crop and zoom in on the relevant area:
{\bfseries\ttfamily \string<code\string>}\par
\lstset{style=modelcode}
\begin{lstlisting}
import os
import random
from PIL import Image

# Load the image
image_path = "48.jpg"
image = Image.open(image_path)

# Define the cropping coordinates (x1, y1, x2, y2)
x1, y1, x2, y2 = 1500, 500, 4000, 3000

# Crop the image
cropped_image = image.crop((x1, y1, x2, y2))

# Save the cropped image with a simple filename
processed_path = "cropped_1.jpg"
cropped_image.save(processed_path)

# Print the path of the saved cropped image
print(processed_path)
\end{lstlisting}
{\bfseries\ttfamily \string</code\string>}
{\bfseries\ttfamily \string</think\string>}\par
\end{thinkcodebox}

\bigskip

\begin{sandboxbox}[Sandbox Output]
\centering
\includegraphics[width=0.4\linewidth]{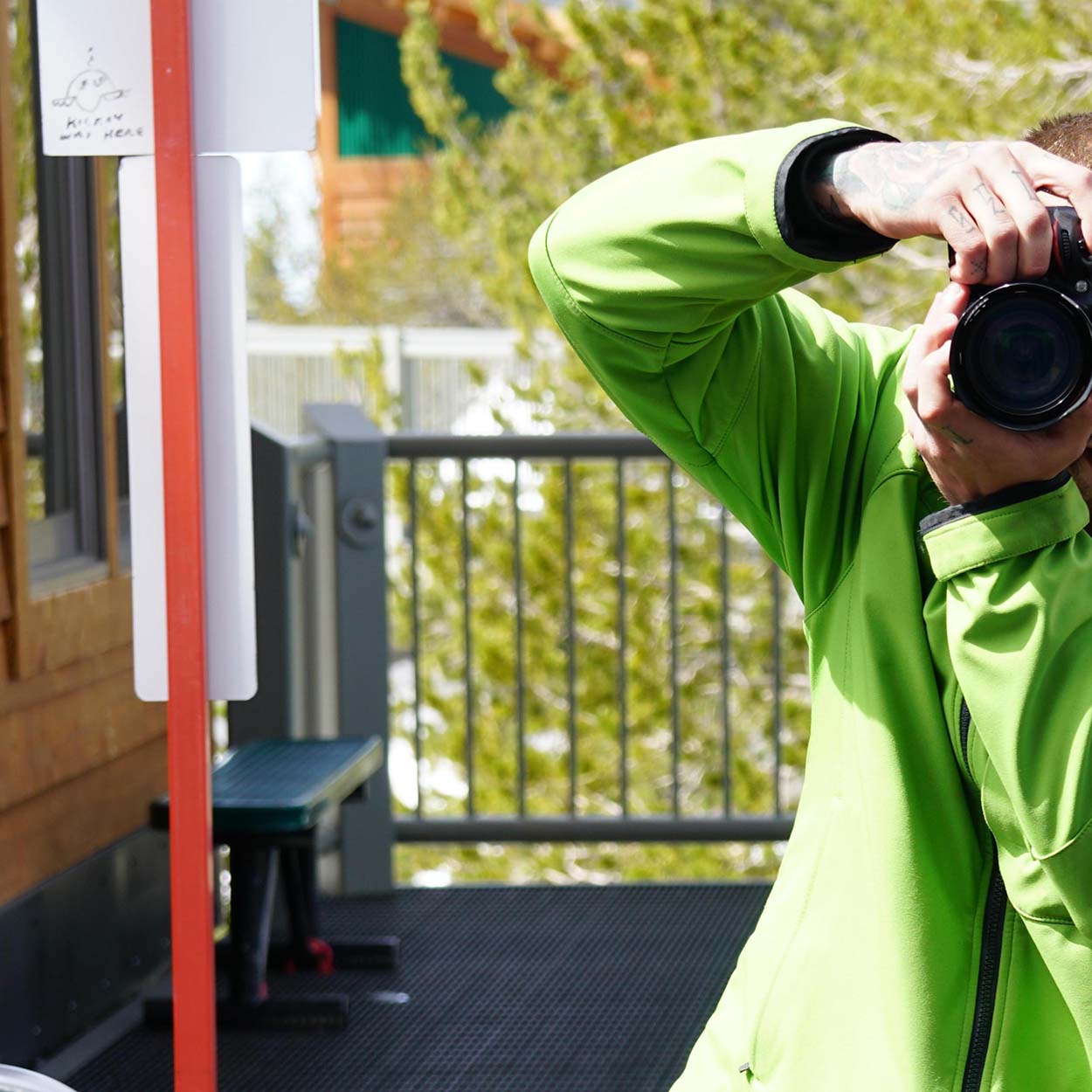}
\end{sandboxbox}

\bigskip

\begin{thinkcodebox}[Assistant]
{\bfseries\ttfamily \string<think\string>}
The initial cropping did not effectively capture the relative position of the person in relation to the recycle bin. The cropped image focused too closely on the person, excluding the recycle bin and the surrounding area. To accurately determine the person's position relative to the recycle bin, the cropping range should be expanded to include both the person and the recycle bin.

The corrected cropping code should adjust the coordinates to ensure both the person and the recycle bin are visible. The person is still in the foreground, but the recycle bin should be included in the background to the right.

Here is the corrected cropping code:

{\bfseries\ttfamily \string<code\string>}\par
\lstset{style=modelcode}
\begin{lstlisting}
import os
import random
from PIL import Image

# Load the image
image_path = "48.jpg"
image = Image.open(image_path)

# Define the corrected cropping coordinates (x1, y1, x2, y2)
x1, y1, x2, y2 = 0, 0, 7000, 4000

# Crop the image
cropped_image = image.crop((x1, y1, x2, y2))

# Save the cropped image with a simple filename
processed_path = "cropped_1.jpg"
cropped_image.save(processed_path)

# Print the path of the saved cropped image
print(processed_path)
\end{lstlisting}
{\bfseries\ttfamily \string</code\string>}
{\bfseries\ttfamily \string</think\string>}\par
\end{thinkcodebox}

\bigskip
\begin{sandboxbox}[Sandbox Output]
\centering
\includegraphics[width=\linewidth]{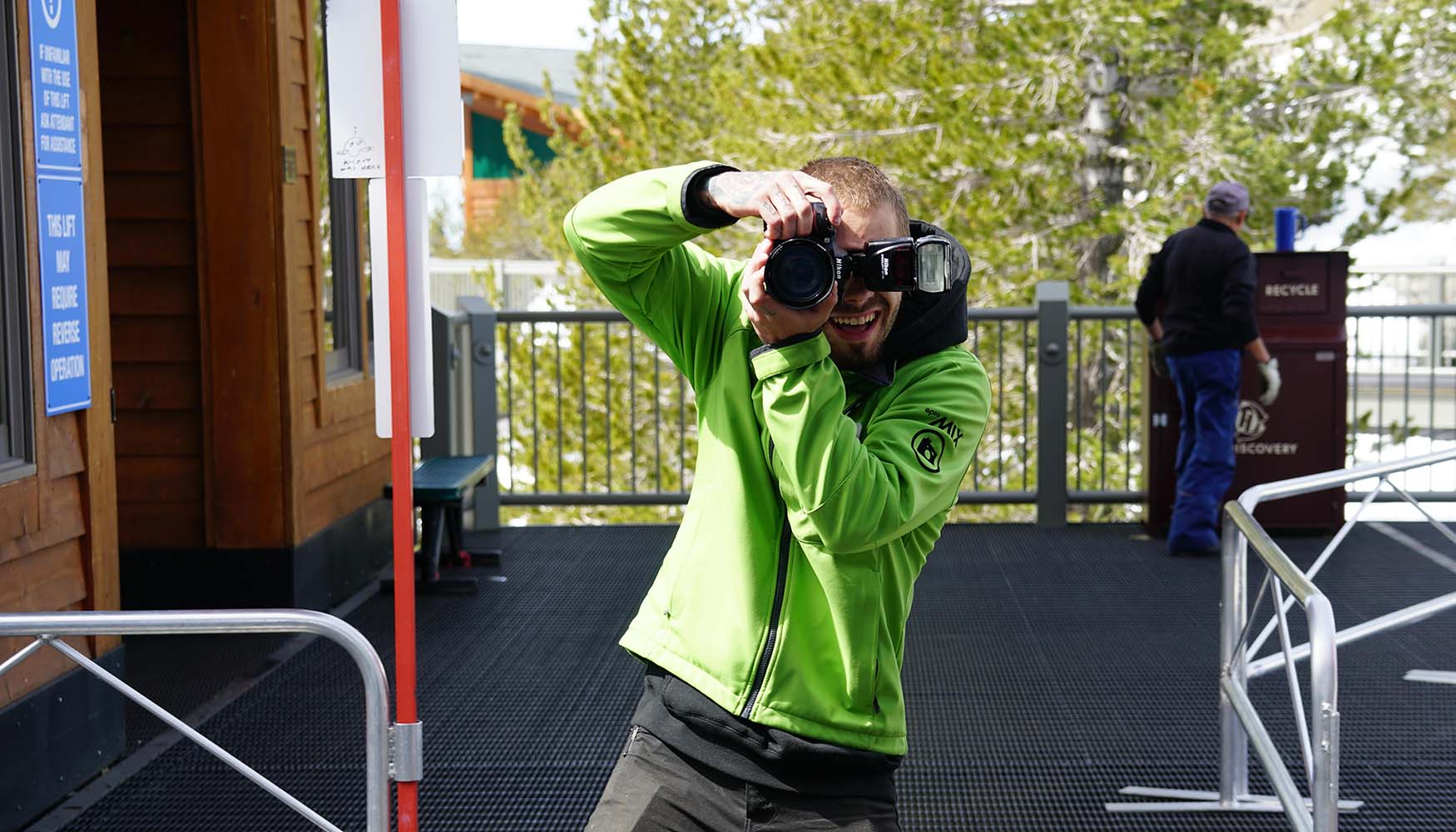}
\end{sandboxbox}
\begin{answerbox}[Assistant]
{\bfseries\ttfamily \string<think\string>}The final cropped image now clearly shows both the person and the recycle bin. The person is standing in front of the recycle bin, which is visible in the background to their right. The person is positioned closer to the camera, while the recycle bin is further back and to the right. This confirms that the person is in front of the recycle bin.
{\bfseries\ttfamily \string</think\string>}

{\bfseries\ttfamily \string<answer\string>}A. In front of the recycle bin{\bfseries\ttfamily \string</answer\string>}
\end{answerbox}

\end{document}